\documentclass[letterpaper]{article} % DO NOT CHANGE THIS
\usepackage[]{aaai2026}  % DO NOT CHANGE THIS
\usepackage{times}  % DO NOT CHANGE THIS
\usepackage{helvet}  % DO NOT CHANGE THIS
\usepackage{courier}  % DO NOT CHANGE THIS
\usepackage[hyphens]{url}  % DO NOT CHANGE THIS
\usepackage{graphicx} % DO NOT CHANGE THIS
\usepackage{natbib}  % DO NOT CHANGE THIS AND DO NOT ADD ANY OPTIONS TO IT
\usepackage{caption} % DO NOT CHANGE THIS AND DO NOT ADD ANY OPTIONS TO IT
\usepackage{amsmath}
\usepackage{siunitx}
\usepackage{algorithm}
\usepackage{algorithmic}
\usepackage{amsfonts}
\usepackage{multirow}%
\usepackage{multicol}
\usepackage{colortbl}
\usepackage{xcolor}
\usepackage{amsthm}%
\usepackage{amsmath,amssymb,amsfonts}%
\usepackage{tabularx}
\usepackage{float}
\usepackage{makecell}
\usepackage{diagbox}
\usepackage{times}
\usepackage{arydshln} 
\usepackage{booktabs}
\usepackage{newfloat}
\usepackage{listings}

\usepackage[most]{tcolorbox}
\usepackage{pifont}          % 提供 \cmark 和 \xmark 等符号
\usepackage{textcase}        % 提供 \textsc{} 命令（大多数文类默认支持）

\nocopyright

\DeclareCaptionStyle{ruled}{labelfont=normalfont,labelsep=colon,strut=off} % DO NOT CHANGE THIS
\floatstyle{ruled}
\newfloat{listing}{tb}{lst}{}
\floatname{listing}{Listing}
\title{MemGuide: Intent-Driven Memory Selection for Goal-Oriented \\ Multi-Session LLM Agents}

\author {
    Yiming Du\textsuperscript{\rm 1}\thanks{These authors contributed equally.},
    Bingbing Wang\textsuperscript{\rm 2}\footnotemark[1],
    Yang He\textsuperscript{\rm 3},
    Bin Liang\textsuperscript{\rm 1},
    Baojun Wang\textsuperscript{\rm 4},\\
    Zhongyang Li\textsuperscript{\rm 4},
    Lin Gui\textsuperscript{\rm 5},
    Jeff Z. Pan\textsuperscript{\rm 6},
    Ruifeng Xu\textsuperscript{\rm 2},
    Kam-Fai Wong\textsuperscript{\rm 1}
}
\affiliations {
    \textsuperscript{\rm 1}The Chinese University of Hong Kong,
    \textsuperscript{\rm 2}Harbin Institute of Technology, Shenzhen\\
    \textsuperscript{\rm 3}The Hong Kong University of Science and Technology,
    \textsuperscript{\rm 4}Huawei Noah’s Ark Lab,
    \textsuperscript{\rm 5}King’s College London,\\
    \textsuperscript{\rm 6}The University of Edinburgh,
    \{ydu, kfwong\}@se.cuhk.edu.hk, bingbing.wang@stu.hit.edu.cn
}

\usepackage{bibentry}
\begin{document}

\maketitle

\begin{abstract}
Modern task-oriented dialogue (TOD) systems increasingly rely on large language model (LLM) agents, leveraging Retrieval-Augmented Generation (RAG) and long-context capabilities for long-term memory utilization. However, these methods are primarily based on semantic similarity, overlooking task intent and reducing task coherence in multi-session dialogues. To address this challenge, we introduce \textbf{MemGuide}, a two-stage framework for intent-driven memory selection. (1) \textbf{Intent‑Aligned Retrieval} matches the current dialogue context with stored intent descriptions in the memory bank, retrieving QA‑formatted memory units that share the same goal. (2) \textbf{Missing‑Slot Guided Filtering} employs a chain‑of‑thought slot reasoner to enumerate unfilled slots, then uses a fine‑tuned LLaMA‑8B filter to re‑rank the retrieved units by marginal slot‑completion gain. The resulting memory units inform a proactive strategy that minimizes conversational turns by directly addressing information gaps. Based on this framework, we introduce the \textbf{MS-TOD}\footnote{Code and dataset will be released upon paper acceptance.}, the first multi-session TOD benchmark comprising 132 diverse personas, 956 task goals, and annotated intent-aligned memory targets, supporting efficient multi‑session task completion. Evaluations on MS-TOD show that MemGuide raises the task success rate by 11\% (88\%→99\%) and reduces dialogue length by 2.84 turns in multi-session settings, and maintains parity with  single‑session benchmarks.
\end{abstract}

% \cite{wang2108task, he2022space, bang2023task, swamy2023contextual} have traditionally focused on single-session scenarios, overlooking the fact that real-world interactions often span multiple sessions over extended periods. While LLMs have been introduced to improve TOD \cite{xu2024rethinking, xu2024hr, chung2023instructtods, heck2023chatgpt}, most efforts remain confined to single-session settings and overlook long-term memory augmentation across multi-session interactions \citep{du2025rethinking}.
\begin{figure}[!t]
  \centering
  \includegraphics[width=\linewidth]{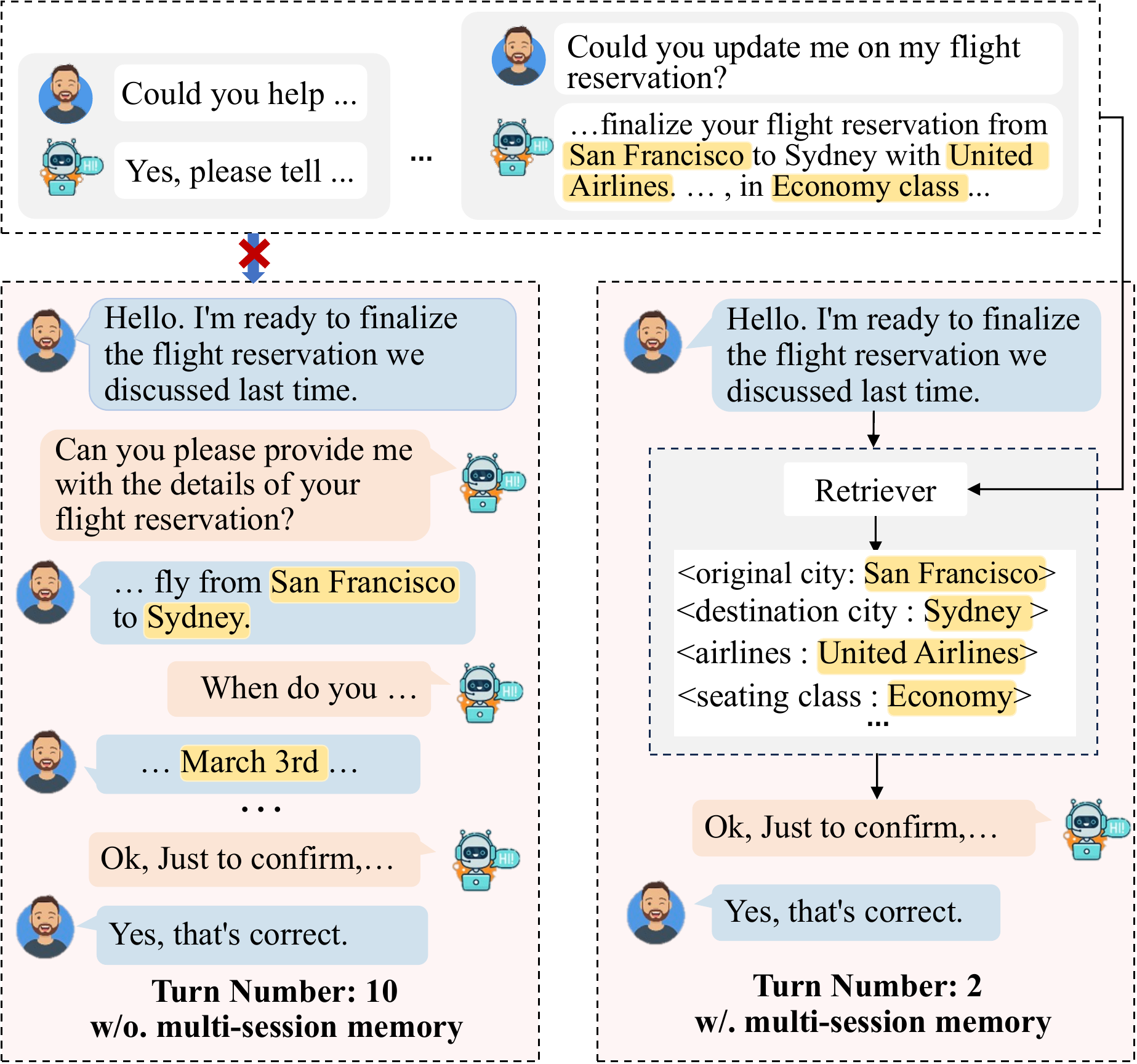}
  \caption{Task-oriented dialogue, without (\textit{left}) vs. with (\textit{right}) multi-session memory; the former demands more turns of conversation.}
  \label{f-intro-example}
\end{figure}

\section{Introduction}
Modern task-oriented dialogue (TOD) systems increasingly integrate large language models (LLMs) to enhance generalization, context understanding, and response generation \citep{nguyen2025spec, xu2024rethinking, xu2024hr, chung2023instructtods, hudecek-dusek-2023-large}. To utilize historical dialogue states across turns, two dominant strategies have emerged: retrieval-augmented generation (RAG), which supplements the LLM with relevant task descriptions or prior dialogue states \citep{xu2024rethinking, xu2024hr}, and the use of long-context models, which encode the entire history directly as input \citep{nguyen2025spec}. However, they are both limited to surface-level semantic similarity, often ignoring the task-specific intent and slot-level continuity crucial for coherent multi-session TOD. 

While memory-augmented methods \citep{lu2023memochat} have emerged, they are often evaluated on single-session benchmarks that lack support for long-term goal tracking and multi-session memory supervision. Unlike open-domain settings that focus on free-form recall \citep{zhong2024memorybank}, task-oriented dialogue demands structured slot tracking, evolving intent management, and consistent state maintenance. Crucially, real-world users frequently interact with assistants across multiple sessions to accomplish complex goals, yet most existing TOD models and datasets \citep{budzianowski2018multiwoz, rastogi2020towards, stacey2024lucid, liu2024toad} are confined to single-session settings, highlighting a fundamental gap between current TOD systems and the demands of persistent, goal-driven dialogue settings. As shown in Figure~\ref{f-intro-example}, traditional single-session TOD systems require users to restate details (e.g., flight times, seat preferences) in every session, leading to inefficiency and frustration.

% To address this, multi-session memory enables retrieval of prior user-specific information, such as booking history, thus reducing redundancy and enabling more personalized and efficient goal completion. Therefore, 

To address this, we propose \textbf{MemGuide}, a novel framework that utilizes long-term memory in the multi-session TOD task. Unlike prior methods that rely solely on semantic similarity for memory selection, MemGuide incorporates task intent and slot-level guidance to enhance memory relevance and response quality. It consists of two core phases: (1) \textbf{Intent-Aligned Retrieval}, where an LLM generates an intent hypothesis based on the current dialogue state and retrieves memory units aligned with the predicted goal, ensuring retrieved history is both semantically similar and goal-consistent to support cross-session task tracking robustly. (2) \textbf{Missing-Slot Guided Filtering}, which first employs a chain-of-thought (CoT) slot reasoner to detect missing slot values, followed by a filtering module that removes irrelevant or redundant Question Answering (QA) memory units to distill slot-level content for response generation. 
% This distillation process provides precise slot-level supervision and ensures that the system reuses only goal-relevant history. 
These two phases enable MemGuide to convert long-term user history into actionable context, supporting minimal-turn, task-consistent dialogue generation across sessions.

To enable systematic evaluation, we construct the \textbf{Multi-Session Task-oriented Dialogue (MS-TOD)}, a new benchmark comprising 132 simulated speakers, each engaging in over 20 sessions covering diverse task goals derived from Schema-Guided Dialogue (SGD) \cite{rastogi2020towards}. MS-TOD supports evaluation of slot continuity and long-term memory retrieval across sessions. Unlike open-domain benchmarks focused on retrieving dialogue summaries \cite{zhong2024memorybank, li2024hello, du-etal-2024-perltqa}, multi-session TOD introduces additional challenges. Systems must recall key slot-value pairs, track evolving user intents, and proactively resolve missing or outdated information, while minimizing redundant interactions. To support automatic evaluation, we introduce a \textbf{proactive response generation module} to simulate user engagement and evaluate system performance in resolving missing information. Experimental results demonstrate that MemGuide significantly improves dialogue coherence, response quality, task success rate, and overall efficiency in multi-session TOD.
The main contributions include:
\begin{itemize}
    \item We propose \textbf{MemGuide}, a two-stage framework that distills and leverages cross-session memory for efficient, minimal-turn task completion.
    \item We introduce \textbf{MS-TOD}, the first multi-session TOD dataset and benchmark task for evaluating long-term memory integration across sessions.
    \item We demonstrate that MemGuide consistently outperforms strong baselines across multiple metrics, demonstrating the effectiveness of intent-aware memory retrieval and slot-guided filtering.
\end{itemize}

% To ensure end-to-end consistency, we further introduce a \textbf{proactive response generation module} that detects slot mismatches via response–goal alignment and actively engages users to complete missing information, thereby reducing redundancy and enhancing task success.

\section{Related Work}
\subsection{Task-Oriented Dialogue Dataset}
TOD datasets are typically constructed via either Machine-to-Machine (M2M) \cite{shah2018building, rastogi2020towards} or Wizard-of-Oz (WOz) setups \cite{wen2017network, budzianowski2018multiwoz}. M2M datasets (e.g., SGD, STAR) provide schema-driven task flows, while WOz-based datasets (e.g., MultiWOZ, FRAMES) offer more natural but annotation-heavy dialogues. Recent efforts aim to improve realism and domain diversity \cite{hu2023multi, dai2022cgodial, xu2024hr, li2024opera}, yet existing benchmarks primarily assume single-session tasks. There remains a notable gap in datasets designed for multi-session TOD, where tracking long-range goals and user intents is essential.

\subsection{Task-Oriented Dialogue Systems}
Traditional TOD systems adopt modular pipelines for NLU, DST, and response generation \cite{wu2019transferable, peng2018deep}, later unified into end-to-end models trained on annotated dialogues \cite{wen2017network, wang2020multi}. With the rise of LLMs, recent work explores their use in zero-shot and fine-tuned TOD \cite{madotto2021few, bang2023task}, often achieving strong results on intent recognition and slot filling. In parallel, long-term memory (LTM) methods such as ChitChat \cite{li2024chatcite}, MemoryBank \citep{zhong2024memorybank}, and LoCoMo \cite{maharana2024evaluating} support extended context retention through summarization or heuristic filtering, but lack structured memory aligned with task goals. Most assume single-session dialogues and overlook challenges in maintaining multi-session goal continuity. This work addresses these gaps by introducing MemGuide for long-range, goal-aware tracking.

\begin{table}[t]
    \centering
    \renewcommand{\arraystretch}{1} % 增加行间距
    \setlength{\tabcolsep}{4pt} % 调整列间距
    \small % 适配单栏排版

    \begin{tabular}{p{3.8cm} c c} 
        \toprule
        \textbf{Settings} & \textbf{GPT-4 Score} & \textbf{Slot Acc.} \\
        \midrule
        \multicolumn{3}{l}{\cellcolor[HTML]{EFEFEF}\textbf{No Retrieval (Direct Prompting)}} \\  
        \makecell[l]{Current Session Context}  & 2.60  & 0.13 \\ 
        \makecell[l]{Full Context} & 4.76 & 0.61 \\ 
        \cdashline{1-3} % 添加虚线分隔
        
        \multicolumn{3}{l}{\cellcolor[HTML]{EFEFEF}\textbf{Retrieval-based Methods}} \\  
        \makecell[l]{BM25-Based Retrieval}   & 5.90  & 0.53 \\ 
        \makecell[l]{Embedding-Based Retrieval}  & 7.01  & 0.67 \\ 
        \makecell[l]{Hybrid Retrieval}     & 7.04  & 0.68 \\ 
        \cdashline{1-3} % 添加虚线分隔
        
        \multicolumn{3}{l}{\cellcolor[HTML]{EFEFEF}\textbf{Oracle (Upper Bound)}} \\  
        \makecell[l]{Oracle}    & \textbf{8.51}  & \textbf{0.82} \\ 
        \bottomrule
    \end{tabular}

    \caption{Evaluation of confirmation-type response generation under different prompting and retrieval strategies.}
    \label{tab:retrieval_generation}
\end{table}

\section{Preliminary Experiments}\label{sec:preliminary exp}

To motivate our framework, we first examine the limitations of direct prompting in multi-session TOD and explore the potential of retrieval-based strategies. Since existing TOD datasets lack long-term dependencies, we construct an evaluation set focused on \emph{confirmation-type} response generation. This evaluation set adopts the same input–output formulation, evaluation criteria, and sample set as the MS-TOD benchmark to ensure consistency and comparability.

To clarify the evaluation setting, we formulate a confirmation-style response generation task. Formally, given the current user utterance $u_t$, the dialogue context $c = \{u_1, r_1, \dots, u_{t-1}, r_{t-1}\}$ from the current session, and history dialogue sessions $H$ in the corresponding memory bank (retrieved or concatenated), the model generates a confirmation-style response $r_t$. The task goal $g$ is held out for reference during evaluation. This setting mimics a common scenario in TOD where the system determines whether the dialogue contains sufficient information to proceed with task execution. For all settings, we use GPT-4o-mini as the unified generator, allowing a fair comparison across input strategies. We compare two strategies:  

% and follows the same input–output formulation and evaluation criteria as our MS-TOD benchmark
(1) \textbf{Retrieval-based methods}, including sparse (BM25 \cite{robertson2009probabilistic}), dense (text-embedding-small-3\footnote{OpenAI. text-embedding-3-small. 2025. \url{https://platform.openai.com/docs/guides/embeddings}}), and a hybrid retrieval. Each selects top-$k$ history sessions $H$ relevant to the current utterance $u_t$ with the model input $x = [H; c; u_t]$. 
(2) \textbf{Direct prompting}, where the full dialogue history is concatenated with the current user utterance $u_t$ without retrieval. The input is $x = [H; C; u_t]$. The model generates a confirmation response $r_t = \text{LLM}(x)$.

% As shown in Table \ref{tab:retrieval_generation}, Retrieval-based methods consistently outperform direct prompting. For instance, \emph{dense retrieval}-based method achieves 0.67 slot accuracy and 7.01 GPT-4 score, surpassing full-context prompting (0.61 and 4.76). \emph{Hybrid retrieval} improves slot accuracy to 0.68 and GPT-4 score to 7.04, demonstrating the value in combining sparse and dense strategies. Oracle retrieval (using ground-truth context) reaches 0.88 and 8.51, underscoring the need for more accurate retrieval strategies in multi-session TOD.

% As Table \ref{tab:retrieval_generation} shows, retrieval-based methods consistently outperform direct prompting. This significant gap highlights how direct prompting struggles with context window limitations and the "lost-in-the-middle" \cite{liu2023lost}, where irrelevant dialogue history overwhelms key information. In contrast, retrieval acts as a filter, providing the LLM with a focused set of relevant sessions. This cleaner input enables more accurate decisions and improves slot accuracy and response quality. This result strongly motivates our development of MemGuide, a new framework using advanced intent-aligned retrieval to overcome these limitations.

As Table \ref{tab:retrieval_generation} shows, retrieval-based methods consistently outperform direct prompting. For instance, \emph{dense retrieval}-based method achieves 0.67 slot accuracy and 7.01 GPT-4 score, surpassing full-context prompting (0.61 and 4.76). This significant gap highlights how direct prompting struggles with context limitations and the "lost-in-the-middle" \cite{liu2023lost}, where irrelevant history overwhelms key information, motivating our development of MemGuide for advanced intent-aligned retrieval.

\begin{figure}[t!]
  \centering
  \includegraphics[width=\linewidth]{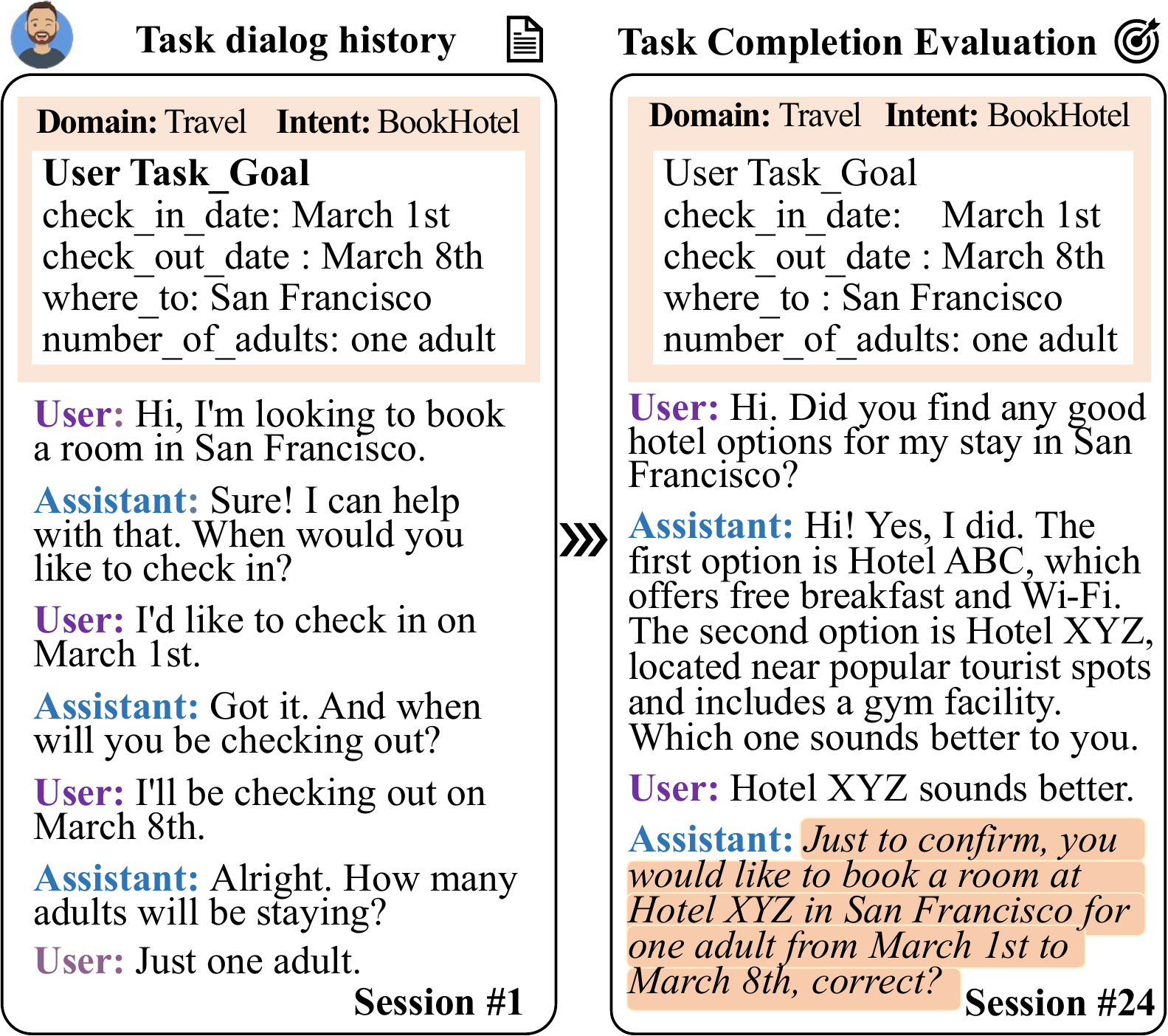}
  \caption{An Example of MS-TOD dataset.}
  \label{f-example}
\end{figure}

\section{Dataset}
\label{sec:dataset}
MS-TOD is a multi-session benchmark for evaluating long-term memory through user-specific memory banks, with over 20 sessions from a single user. This supports assessment of memory retrieval, slot tracking, and intent continuity. For evaluation, we include held-out sessions with manually annotated confirmation-type responses (Figure~\ref{f-example}). MS-TOD is built in four stages: (1) multi-session dialogue generation; (2) confirmation-type response annotation; (3) QA-style memory bank construction; and (4) human validation.

\subsection{Multi-Session Dialogue Generation} 
We begin by generating multi-session dialogues for each task goal sampled from the SGD dataset~\cite{rastogi2020towards}. For every task, we synthesize \emph{three} temporally ordered sessions using GPT-4, each conditioned on the slot-filling status of the previous session. This simulates how users revisit and revise the same task across time. Specifically, Session 1 presents an incomplete task with missing slots; Session 2 introduces updates; and Session 3 concludes with final confirmation. This staged construction reflects real-world dialogue dynamics while avoiding overfitting, as most SGD tasks involve fewer than ten slots.
% More details can be found in Appendix  \ref{app:dataset_generation_prompt}.

\subsection{Confirmation-Type Response Annotation} 
To evaluate long-term task fulfillment in dialogue systems, we annotate the final session of each task with \textit{confirmation-type responses}. Each marks the utterance confirming task completion and associated slot-value goal with manually labels (confirmation/non-confirmation). These annotations serve two purposes: (1) \textbf{Supervising Memory Selection}, indicating when to trigger memory retrieval; and (2) \textbf{Supporting Evaluation}, evaluating if the system recalls goal-relevant content and generates accurate confirmations.

\begin{table}[t!]
\centering

\renewcommand{\arraystretch}{1} % 调整行间距，1 是默认间距

\begin{tabular}{@{}lc@{}}
\toprule
\textbf{Attribute}            & \textbf{Evaluation} \\ \midrule
Domains                      & 16           \\
Intents                      & 19           \\
Task goals                   & 956          \\
Dialogues                    & 2,861        \\
Utterances                   & 18,530       \\
Avg. slots per task goal     & 4.24         \\ \midrule
Number of individuals        & 132          \\
Avg. intents per individual  & 5.45         \\
Avg. sessions per individual & 21.67        \\
Avg. utterances per individual & 140.38     \\ \bottomrule
\end{tabular}

\caption{MS-TOD dataset statistics for evaluation.}
\label{tab:dataset_statistic_new}
\end{table}

\begin{figure*}[htbp]
  \centering
  \includegraphics[width=0.85\linewidth]{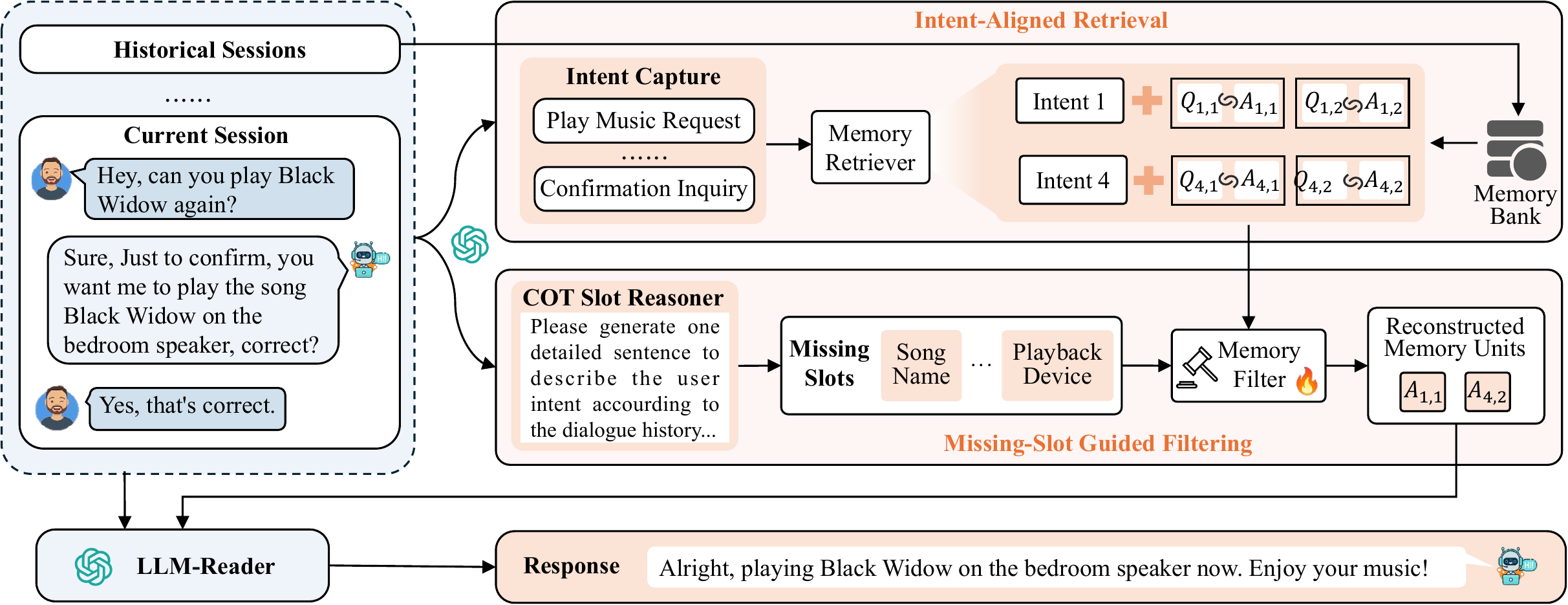}
  \caption{Overflow of our MemGuide framework, which comprises Intent-Aligned Retrieval and Missing-Slot Guided Filtering.}
  \label{f-framework}
\end{figure*}

\subsection{Memory Bank Construction}
Since multi-session interactions are organized around individuals, we group dialogues into \textit{individual memory banks} (Figure~\ref{f-example}), where each bank stores temporally ordered sessions. During construction, we ensure that intents within the same bank are distinct and non-conflicting, enabling consistent memory usage and avoiding cross-intent interference. Each bank contains over 20 sessions spanning at least six task intents (Table~\ref{tab:dataset_statistic_new}), with one held-out evaluation session per intent for confirmation-type assessment.  For each completed session, we use GPT-4 to generate an intent description along with a set of QA pairs, each capturing a slot-specific fact for retrieval. This QA-style format is motivated by prior work \citep{chen-etal-2023-augmenting} showing that question–answer structures facilitate more accurate and efficient retrieval compared to unstructured text. To maintain causal consistency, memory access is limited to sessions prior to the current evaluation point.
% (Appendix~\ref{sec:dataset_design}

\subsection{Human Validation} 
To ensure coherence, correctness, and usability, we apply a structured multi-stage validation process involving three annotators experienced in NLP. The process includes: (1) Verifying  each session preserves the intended goal and correct slot values; (2) Removing dialogues with excessive redundancy across sessions; (3) Verifying that confirmation-type utterances match the expected slot-value goals; (4) Removing sessions that fail to complete defined task goals or lack confirmation turns; and (5) Excluding episodes with unnatural repetition of similar intents. We additionally conduct inter-annotator agreement evaluation to ensure labeling consistency.

\section{MemGuide}

MemGuide is a two-stage framework for multi-session TOD, as shown in Figure~\ref{f-framework}. It first performs \textbf{\textit{Intent-Aligned Retrieval}}, extracting the current user intent via an LLM and retrieving QA-formatted memory units from the memory bank that align with this intent using semantic similarity. Then, \textbf{\textit{Missing-Slot Guided Filtering}} identifies unfilled task slots through a CoT reasoner and re-ranks retrieved memories with a fine-tuned LLaMA-8B filter based on their ability to fill these slots. Finally, \textbf{\textit{Response Generation}} leverages the filtered memories to produce proactive responses, reducing conversational turns and boosting task success.

% \subsection{Task Definition}
Given a dialogue context $c = \{u_1, r_1, \dots, u_{t-1}, r_{t-1}, u_{t}\}$ and a user-specific long-term memory bank $M$, the goal is to generate the next system response $r_{t}$. The memory bank $M$ is a collection of memories from past sessions, structured as  $M = \{ (k_i, V_i) \}_{i=1}^{N}$ and $t$ is the number of turns. Each entry consists of a high-level intent description $k_i$ (e.g., "book a flight to San Francisco") represents the high-level intent and a corresponding set of QA-formatted memory units $V_i = \{ (q_{i,j}, a_{i,j}) \}_{j=1}^{n}$, where $q_{i,j}$ is the $j$-th question about a task slot within the $i$-th intent (e.g., "What is the departure date?") and $a_{i,j}$ is the corresponding answer (e.g., "July 28, 2025").
The optimal response $r$ should coherently continue the conversation while strategically utilizing information from $M$  to progress towards task completion. It is generated by an LLM conditioned on the context and a carefully selected memory subset  $M_{sel} \subseteq M$.

\subsection{Intent-Aligned Retrieval} 
The stage aims to broadly identify past sessions relevant to the current user goal by retrieving memory entries from the long-term bank that share a consistent high-level intent, ensuring thematic alignment.

\subsubsection{Current Intent Extraction.} Given the dialogue context $c$, we use GPT-4o-mini to generate a high-level intent description \( d_{int} \), which summarizes the user's objective in the current session. The model is prompted to generate a short, command-like phrase summarizing the user's goal (e.g., ''find a local Italian restaurant''). This distilled phrase serves as a current intent key, $k_{cur}$, providing a canonical task representation that is standardized for retrieval.

\subsubsection{Semantic Retrieval.}
We then retrieve memory units from the bank $M$, which contain stored intent keys $ \{k_i\}_{i=1}^{N}.$ and corresponding QA pairs that are semantically closest to the extracted current intent $k_{cur}$. We use an embedding model (e.g., text-embedding-3-small) to compute dense vector representations for all memory units. The relevance score between $k_{cur}$ and each memory unit is calculated using cosine similarity. The top-$K$ memory units $(k_i, V_i)$ with the highest scores are selected to form the candidate memory set, $M_{cand}$. This procedure ensures that we only consider memories from sessions with a shared high-level objective, guaranteeing thematic alignment for subsequent stages. 

\subsection{Missing-Slot Guided Filtering.} 
While the retrieved memories in $M_{cand}$ are thematically aligned with the user's intent, their immediate utility for the current dialogue turn can vary significantly. To address this, this stage performs a fine-grained filtering, prioritizing memory units that are most likely to resolve the immediate information needs of the dialogue.

\subsubsection{Information Gap Identification.} To guide the filtering process, we first must precisely identify what information is required to advance the current task. 
We leverage an LLM configured as a CoT reasoner \cite{wei2022chain} to analyze the dialogue context $c$ and the overall goal corresponding with intent key $k_{cur}$, so as to enumerate a list of essential task slots that have not yet been filled or confirmed. 
The CoT prompt is structured to guide the LLM through a logical sequence.
First, it enumerates all required slots for the intent, then checks the dialogue history to determine which of these slots have already been filled or confirmed, and finally outputs only those slots that remain unresolved. The result is a list of hypothesized missing slots, $L_{miss} = \{slot_1, slot_2, \dots\}$. For example,  in a flight booking task where the user has only specified a destination, $L_{miss}$ might be identified as $\{\text{departure\ date}, \text{return\ date}, \text{seat\ preference}\}$.

\subsubsection{Re-ranking by Marginal Slot-Completion Gain.}
Having identified the information gaps represented by $L_{miss}$, we then re-rank each QA pair \((q_{i,j}, a_{i,j})\) within the candidate set $M_{cand}$ based on its potential to fill one of these gaps. This process is driven by the principle of selecting information with the highest marginal slot-completion gain. 

To operationalize this, we fine-tune a smaller, efficient LLaMA-8B model \citep{meta2024llama3.1} to act as a specialized filter. This filter estimates the probability that a given QA pair provides an answer for one of the missing slots. For each QA pair \((q_{i,j}, a_{i,j})\) from $M_{cand}$, the model computes:
\begin{equation}
s_{i,j} = P(y=1 \mid c, L_{\text{miss}}, q_{i,j}, a_{i,j})
\end{equation}
where $y=1$ signifies that the answer $a_{i,j}$ successfully fills a slot presented in $L_{miss}$. To supervise this filter, we construct a training dataset using the same pipeline as MS-TOD memory bank generation. Specifically, for each held-out session, we simulate missing-slot contexts and label each QA pair from prior sessions as positive (if it fills a missing slot) or negative (otherwise). Detailed training configurations and dataset  are provided in Appendix.
The model is optimized using a standard binary cross-entropy loss function:
\begin{equation}
\mathcal{L} = -\sum_{i,j} \left[ y_{i,j} \log s_{i,j} + (1-y_{i,j}) \log (1 - s_{i,j}) \right]
\end{equation}

To balance the initial semantic relevance from the initial semantic retrieval (denoted as $s^{\text{pre}}_{i,j}$) with the slot-filling utility score $s_{i,j}$ from our filter, we compute a final score:
\begin{equation}
s_{\text{final},ij} = \alpha \cdot s^{\text{pre}}_{i,j} + (1 - \alpha) \cdot s_{i,j}
\end{equation}
where $\alpha$ is a hyperparameter. We select the top-$K$ (e,g,. K=5) QA pairs with the highest $s_{\text{final},ij}$ scores.

\subsection{Response Generation}
The final response is generated using the top-$K$ memorys $A_{\text{core}} = \{a_1, a_2, \dots, a_K\}$, omitting auxiliary questions $q_{i,j}$. An LLM reader receives the dialogue context $c$, the core facts $A_{\text{core}}$, and missing slots $L_{\text{miss}}$ as prompt inputs.
\begin{equation}
r = \text{LLM}{\text{Reader}}(\text{prompt}(c, A_{core}, L_{\text{miss}}))
\end{equation}
The prompt instructs the model to: (1) continue the conversation naturally, and (2) proactively address $L_{\text{miss}}$ using $A_{\text{core}}$, e.g., by confirming or suggesting stored values. For example, if memory provided a preferred airline, the system might respond, "I see you're flying to Montreal. Last time you flew with Air Canada. Would you like to book with them again?" This proactive strategy, directly informed by our two-stage memory selection, minimizes redundant questions and accelerates task completion. All prompt templates used in our method are provided in Appendix.

\section{Experiments}
To enable fine-grained assessment of long-term memory utilization and task completion, we evaluate MemGuide on sessions from the MS-TOD benchmark that are annotated with \textit{confirmation-type responses}, in which the final utterance explicitly confirms that the user goal has been achieved. Each session is associated with a gold-standard task goal and corresponding slot-value set, enabling precise evaluation using standard metrics of task success and response quality.

\subsection{Experimental Setups}
\subsubsection{Evaluation Metrics.} We use four core automatic metrics and human evaluation to evaluate response performance: 1) \textbf{GPT-4 score}, (1–10) \footnote{GPT-4-as-the-judge prompts can be found in Appendix} evaluates response quality in terms of fluency, coherence, and informativeness;
2) \textbf{Joint Goal Accuracy} (JGA) measures slot prediction accuracy;
3) \textbf{Dialogue Turn Efficiency} (DTE) captures the number of turns required to complete a task, and 
4) \textbf{Success Rate} (S.R.) indicates whether the user goal is achieved. 5) \textbf{Human evaluation} further assesses Accuracy, Informativeness, and Coherence, with \textbf{A.I.C}. denoting their average.
% Annotation details are provided in Appendix~\ref{app: human_evalution}.
% To support analysis, we report auxiliary metrics including 5) \textbf{Recall@k} for memory retrieval accuracy, 
% 6) \textbf{Slot Accuracy} for value correctness, and 
% 7) \textbf{BLEU} and 8) \textbf{ROUGE} for generation overlap. 

\begin{table*}[t!]
\centering
\small
\renewcommand{\arraystretch}{1}
\begin{tabular}{llcccccccc}
\toprule
\textbf{Model} & \textbf{Setting} & \textbf{GPT-4} & \textbf{JGA} & \textbf{DTE} & \textbf{S.R.} & \textbf{A.} & \textbf{I.} & \textbf{C.} & \textbf{A.I.C.} \\
\midrule
\multirow{2}{*}{\textbf{LLaMA3-8B}} 
    & FCP & 4.89  & \textbf{0.64}  & 5.37 & 0.82 & 0.56 & 1.47 & 1.74 & 1.26 \\
    & \cellcolor[gray]{0.9}MemGuide & \cellcolor[gray]{0.9}\textbf{6.39}  & \cellcolor[gray]{0.9}0.63  & \cellcolor[gray]{0.9}\textbf{3.46} & \cellcolor[gray]{0.9}\textbf{0.92} & \cellcolor[gray]{0.9}\textbf{0.61} & \cellcolor[gray]{0.9}\textbf{1.98} & \cellcolor[gray]{0.9}\textbf{2.16} & \cellcolor[gray]{0.9}\textbf{1.58} \\
\hdashline
\multirow{2}{*}{\textbf{Qwen-7B}} 
    & FCP & 6.26  & 0.66  & 4.93 & 0.83 & 0.43 & 1.24 & 1.85 & 1.17 \\
    & \cellcolor[gray]{0.9}MemGuide & \cellcolor[gray]{0.9}\textbf{6.81}  & \cellcolor[gray]{0.9}\textbf{0.66}  & \cellcolor[gray]{0.9}\textbf{4.31} & \cellcolor[gray]{0.9}\textbf{0.87} & \cellcolor[gray]{0.9}\textbf{0.54} & \cellcolor[gray]{0.9}\textbf{1.70} & \cellcolor[gray]{0.9}\textbf{2.30} & \cellcolor[gray]{0.9}\textbf{1.51} \\
\hdashline
\multirow{2}{*}{\textbf{Mistral-7B}} 
    & FCP & 6.20  & 0.73  & 2.52 & 1.00 & 0.58 & 1.63 & 1.99 & 1.40 \\
    & \cellcolor[gray]{0.9}MemGuide & \cellcolor[gray]{0.9}\textbf{6.48}  & \cellcolor[gray]{0.9}\textbf{0.80}  & \cellcolor[gray]{0.9}\textbf{1.21} & \cellcolor[gray]{0.9}\textbf{1.00} & \cellcolor[gray]{0.9}\textbf{0.61} & \cellcolor[gray]{0.9}\textbf{2.06} & \cellcolor[gray]{0.9}\textbf{2.08} & \cellcolor[gray]{0.9}\textbf{1.58} \\
\hdashline
\multirow{2}{*}{\textbf{GPT-4o-mini}} 
    & FCP & 6.93  & 0.67  & 6.03 & 0.88 & 0.62 & 1.83 & 1.90 & 1.78 \\
    & \cellcolor[gray]{0.9}MemGuide & \cellcolor[gray]{0.9}\textbf{7.14}  & \cellcolor[gray]{0.9}\textbf{0.70}  & \cellcolor[gray]{0.9}\textbf{3.19} & \cellcolor[gray]{0.9}\textbf{0.99} & \cellcolor[gray]{0.9}\textbf{0.65} & \cellcolor[gray]{0.9}\textbf{2.38} & \cellcolor[gray]{0.9}\textbf{2.48} & \cellcolor[gray]{0.9}\textbf{2.17} \\
\bottomrule
\end{tabular}
\caption{Combined results comparing FCP and MemGuide across LLMs. GPT-4, JGA, DTE, and S.R. are automatic metrics; A., I., C., and A.I.C. are human evaluation metrics for accuracy, informativeness, coherence, and their average score.}
\label{tab:combined_results}
\end{table*}

% \begin{table}[t!]
% \centering
% \small
% \renewcommand{\arraystretch}{1} % 调整行间距
% \setlength{\tabcolsep}{4pt} % 调整列间距 % 设置字体大小
% \begin{tabular}{llcccc}
% \toprule
% \textbf{Model} & \textbf{Setting} & \textbf{GPT4} & \textbf{JGA} & \textbf{DTE} &\textbf{S.R.}\\
% \midrule
% \multirow{2}{*}{\textbf{LLaMA3-8B}} 
%     & FCP & 4.89  & \textbf{0.64}  & 5.37 & 0.82 \\
%      &  \cellcolor[gray]{0.9} MemGuide & \cellcolor[gray]{0.9}\textbf{6.39}  & \cellcolor[gray]{0.9}0.63  & \cellcolor[gray]{0.9}\textbf{3.46} &  \cellcolor[gray]{0.9}\textbf{0.92} \\ \hdashline
% \multirow{2}{*}{\textbf{Qwen-7B}} 
%     & FCP & 6.26  & 0.66  & 4.93 & 0.83 \\
%      & \cellcolor[gray]{0.9}MemGuide & \cellcolor[gray]{0.9}\textbf{6.81}  & \cellcolor[gray]{0.9}\textbf{0.66}  & \cellcolor[gray]{0.9}\textbf{4.31} & \cellcolor[gray]{0.9}\textbf{0.87} \\ \hdashline
% \multirow{2}{*}{\textbf{Mistral-7B}} 
%     & FCP & 6.20  & 0.73  & 2.52 & 1.00  \\
%      & \cellcolor[gray]{0.9}MemGuide & \cellcolor[gray]{0.9}\textbf{6.48}  & \cellcolor[gray]{0.9}\textbf{0.80}  & \cellcolor[gray]{0.9}\textbf{1.21} & \cellcolor[gray]{0.9}\textbf{1.00}  \\ \hdashline
% \multirow{2}{*}{\textbf{GPT-4o-mini}} 
%     & FCP & 6.93  & 0.67  & 6.03 & 0.88  \\
%      & \cellcolor[gray]{0.9}MemGuide & \cellcolor[gray]{0.9}\textbf{7.14}  & \cellcolor[gray]{0.9}\textbf{0.70}  & \cellcolor[gray]{0.9}\textbf{3.19} & \cellcolor[gray]{0.9}\textbf{0.99}  \\
% \bottomrule
% \end{tabular}
% \caption{Main results comparing FCP and MemGuide for multi-session TOD under various LLMs.}
% \label{tab_comparison_policy}
% \end{table}

\subsubsection{Baselines.} We evaluate MemGuide against three representative categories of baselines:
1) \textbf{General-purpose LLMs.} We assess full-context prompt (FCP)-based dialogue performance using instruction-tuned models including LLaMA3-8B \cite{touvron2024llama3}, Qwen2.5-7B \cite{qwen2024qwen2.5}, Mistral-7B \cite{mistral2024mistral7b}, and GPT-4o-mini \cite{openai2024gpt4}.
2)\textbf{Traditional Task-Oriented Dialogue Systems.} To evaluate MemGuide under structured DST conditions, we include task-specific baselines such as BERT-DST \cite{chao2019bert}, LDST \cite{feng2024ldst}, and AutoTOD \cite{xu2024rethinking}, where the latter incorporates an external memory module for cross-turn goal tracking. While these models were not originally designed for multi-session scenarios, they represent the strongest available TOD pipelines when adapted to this setting.
3) \textbf{Long-term Summarization.} We implement a summarization-based baseline inspired by ChatCite \cite{li2024chatcite}, which condenses session histories into concise summaries used during inference.
% To ensure a fair assessment of generalization, all models are evaluated on a held-out multi-session test set that is excluded from all training processes.

\subsection{Main Results}

\textbf{Comparision with General-purpose LLMs.} Compared to FCP, MemGuide leverages the same underlying LLM as a memory reader to retrieve and utilize relevant long-term memory, yielding substantial improvements across critical metrics. As illustrated in Table~\ref{tab:combined_results}, our approach consistently outperforms FCP across all tested models, demonstrating robust gains in task accuracy, response quality, and interaction efficiency. For example, when using Mistral-7B as the LLM Reader, MemGuide increases JGA from 0.73 to 0.80 and reduces DTE from 2.52 to 1.21, representing a 52\% reduction. LLaMA3-8B achieves the largest improvement in GPT-4 score (from 4.89 to 6.39), while GPT-4o-mini reduces dialogue turns from 6.03 to 3.19, corresponding to a 47.1\% decrease.. Similar gains are observed with Qwen-7B and other models. These consistent gains across models confirm that integrating memory-guided reasoning with the same base model enhances not only task accuracy but also response relevance and interaction fluency, validating the effectiveness of intent-aligned memory selection in multi-session task-oriented dialogue.

% Integrating MemGuide with general-purpose LLMs yields substantial improvements in task efficiency and response quality (Table~\ref{tab_comparison_policy}). As shown in Table~\ref{tab_comparison_policy}, MemGuide leads to notable gains in response quality and task accuracy. For example, applying it to Mistral-7B increases JGA from 0.73 to 0.80 and reduces DTE by 52.0\%. LLaMA3-8B shows the largest gain in GPT-4 score (from 4.89 to 6.39), while GPT-4o-mini achieves a 47.1\% reduction in dialogue turns. Similar trends are observed for Qwen-7B and other models. These findings underscore the benefits of integrating memory-guided reasoning in enhancing task accuracy, response quality, and interaction efficiency.

\begin{table}[t!]
\centering
\small
\renewcommand{\arraystretch}{1}
% \setlength{\tabcolsep}{4pt} 
% \fontsize{9pt}{11pt}\selectfont
\begin{tabular}{lcccc}
\toprule
\textbf{Model} & \textbf{GPT-4} & \textbf{JGA} & \textbf{DTE} & \textbf{S.R.} \\
\midrule
Bert-DST$^*$     & -    & 0.07 & -    & -    \\
LDST$^*$         & -    & 0.23 & -    & -    \\
AutoTOD$^\dag$   & 6.49 & 0.44 & 7.80 & 0.81 \\
ChatCite & 6.59 & 0.660 & 4.71 & 0.84 \\
\rowcolor[gray]{0.9}
MemGuide         & \textbf{7.14} & \textbf{0.70} & \textbf{3.19} & \textbf{0.99} \\
\bottomrule
\end{tabular}
\caption{Results of traditional TOD models, summary-based methods, and MemGuide. Models marked with \(^*\) focus on DST only. $^\dag$ indicates a simplified AutoTOD pipeline.}
\label{tab:comparison_gpt4_jga_dte}
\end{table}

\begin{table}[t!]
\centering
\small
\renewcommand{\arraystretch}{1} % 调整行间距
\setlength{\tabcolsep}{9pt} % 调整列间距
\fontsize{9pt}{11pt}\selectfont % 设置字体大小
\begin{tabular*}{\linewidth}{clccc}
\toprule
\textbf{Dataset} & \textbf{Methods} &\textbf{JGA} & \textbf{AGA}\\
\midrule
\multirow{6}{*}{\textbf{SGD}} 
     & SGD Baseline  & 0.254 & 0.906\\
     & GOLOMB &0.465 &0.750\\
     & SGP-DST &0.722 &0.913\\
     & TS-DST &0.786 &0.956\\
     & LDST & 0.845 & \textbf{0.994}\\
     &  \cellcolor[gray]{0.9} \(\mathrm{MemGuide}^*\) & \cellcolor[gray]{0.9}\textbf{0.846}  & \cellcolor[gray]{0.9}0.965  \\ 
     \hdashline
\multirow{8}{*}{\textbf{MultiWOZ 2.2}}
     & SGD Baseline & 0.420 & -\\
     & TRADE & 0.454 & -\\
     & DS-DST & 0.517 & -\\
     & TripPy & 0.530 & -\\
     & TOATOD & 0.638 & -\\
     & SDP-DST & 0.576 & 0.985\\
     & LDST & 0.607 & \textbf{0.988}\\
     &  \cellcolor[gray]{0.9} \(\mathrm{MemGuide}^*\) & \cellcolor[gray]{0.9}\textbf{0.879}  & \cellcolor[gray]{0.9}0.976  \\ 
\bottomrule
\end{tabular*}
\caption{Results of different methods on SGD and MultiWOZ 2.2. \(\mathrm{MemGuide}^*\) is a single-session variant of MemGuide, where the missing slot guided filtering is disabled while retaining the QA memory.}
\label{QA_ablation}
\end{table}

\textbf{Comparison with Traditional TOD and Summarization Baselines.}
As existing models are not explicitly designed for multi-session TOD, we compare MemGuide with two representative categories: (1) traditional Dialogue State Tracking (DST)-focused models (BERT-DST, LDST, AutoTOD), and (2) a summarization-based approach inspired by ChatCite.  As shown in Table~\ref{tab:comparison_gpt4_jga_dte}, MemGuide consistently outperforms all baselines across key metrics. Compared to AutoTOD, it increases JGA from 0.440 to 0.698 and reduces DTE from 7.80 to 3.19. Against the summarization baseline, MemGuide achieves clear improvements across all metrics: the GPT-4 score increases from 6.59 to 7.14, JGA improves from 0.66 to 0.70, DTE decreases from 4.71 to 3.19, and the success rate rises from 0.84 to 0.99. These results demonstrate the effectiveness of MemGuide in improving both task performance and dialogue efficiency.

\textbf{Human Evaluation.} We conduct human evaluation to assess the effectiveness of MemGuide in generating confirmation-type responses after memory-guided dialogue planning. Human annotators rate each response along three dimensions: \textbf{Accuracy} (binary), \textbf{Informativeness} (scored from 0 to 3), and \textbf{Coherence} (scored from 0 to 3). The average of these scores, denoted as \textbf{A.I.C.}, provides an overall measure of perceived response quality. As shown in Table~\ref{tab:comparison_gpt4_jga_dte}, MemGuide consistently improves human-judged quality across all metrics. All evaluations are conducted under a blind review protocol.

\textbf{Generalization to Single-Session DST Tasks.} 
To assess the generalization of MemGuide in single-session settings, we focus on dialogue state tracking (DST), a core task that supports downstream components like policy planning and response generation. DST is a widely used and well-defined task that depends on context understanding and supports downstream dialogue components. We evaluate it on two widely-used single-session DST benchmarks, SGD and MultiWOZ2.2. While both benchmarks focus on DST, differences in annotation schemes and domain coverage result in distinct baseline configurations across SGD and MultiWOZ2.2 (Table~\ref{QA_ablation}). On SGD, MemGuide achieves a state-of-the-art JGA of 0.846, surpassing strong baselines such as LDST \cite{feng2023llmdrivendialoguestatetracking}, GOLOMB \cite{gulyaev2020goalorientedmultitaskbertbaseddialogue}, SGP-DST \cite{ruan2020finetuningbertschemaguidedzeroshot}, and TS-DST \cite{9892082}, and performs comparably to LDST on Average Goal Accuracy (AGA) \cite{rastogi2020towards}. On MultiWOZ2.2, MemGuide* attains a JGA of 0.879, significantly outperforming prior models including TRADE \cite{wu2019transferablemultidomainstategenerator}, TripPy \cite{heck2020trippytriplecopystrategy}, and SDP-DST \cite{lee2021dialoguestatetrackinglanguage}. We attribute the superior performance to QA memory’s ability to capture slot dependencies more effectively in smaller domain settings, confirming its adaptability and robustness across datasets.

\begin{table}[t!]
\centering
\small
\renewcommand{\arraystretch}{1.0}
\begin{tabular}{lcc}
\toprule
\textbf{Setting} & \textbf{w/ Raw History} & \textbf{w/ Intent-QA Memory} \\
\midrule
LLaMA3-8B   & 5.09 & \textbf{6.34} \\
Qwen-7B     & 6.38 & \textbf{6.56} \\
Mistral-7B  & 5.86 & \textbf{6.71} \\
GPT-4o-mini & 7.01 & \textbf{7.14} \\
\bottomrule
\end{tabular}
\caption{Comparison of GPT-4 scores using retrieved history vs. intent-QA memory across different LLM settings.}
\label{tab:intent_qa_comparison}
\end{table}

%Dialog Turn Efficiency DTE
\subsection{Ablation Study}

% We conduct ablations to evaluate contributions of key MemGuide components, including the missing-slot guided filtering module.

\textbf{Effect of intent-aligned Retrieval.} Table~\ref{tab:intent_qa_comparison} shows that our intent-aligned retrieval with structured QA memory consistently improves response quality, outperforming unstructured baselines by up to 1.29 points (e.g., LLaMa3-8B: 5.05 → 6.34). We attribute this enhancement to the alignment between the retrieved structured context and the model's generative reasoning process. By providing intent-anchored QA pairs, the system can reason over task-relevant content with greater precision and mitigated ambiguity, thereby corroborating the theoretical premise that structured supervision is key to enhancing long-context utility in dialogue systems.

\begin{figure}[t!]
  \centering
  \includegraphics[width=\linewidth]{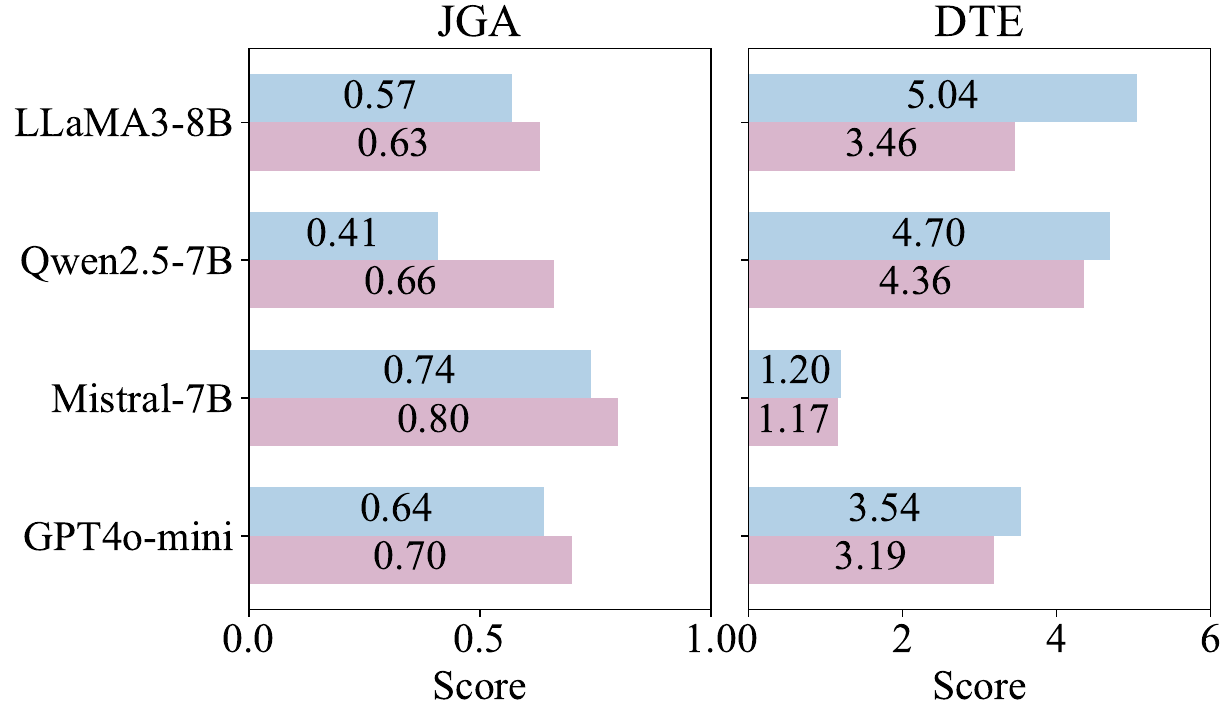}
  \caption{Comparison of Hybrid RAG (blue) vs. MemGuide (purple) across four LLMs on two metrics: JGA and DTE.}
  \label{f-ablation_study_JGA_DTE}
\end{figure}

\textbf{Effect of Missing-Slot Guided Filtering.} We assess the impact of the missing-slot guided filtering module by removing it while retaining the same hybrid RAG retrieval. As shown in Figure~\ref{f-ablation_study_JGA_DTE}, the absence of filtering results in a significant performance drop: JGA on Qwen2.5-7B drops from 0.74 to 0.41, and DTE on GPT-4o-mini increases from 3.19 to 4.30. This highlights the critical role of fine-grained memory selection in both accuracy and interaction efficiency. This module operates in two stages: the CoT reasoner identifies missing task slots, which guide the memory filter to prioritize QA pairs that fill these gaps. This joint reasoning and filtering strategy significantly improves retrieval quality: compared to semantic-only retrieval, Recall@5 improves by 7.7\% on average, raising the performance of text-embedding-3-small from 0.792 to 0.832. By bridging retrieval with task-specific gaps, the filter prioritizes slot-value pairs that are more likely to advance the dialogue.

\begin{figure}[t!]
  \centering
  \includegraphics[width=\linewidth]{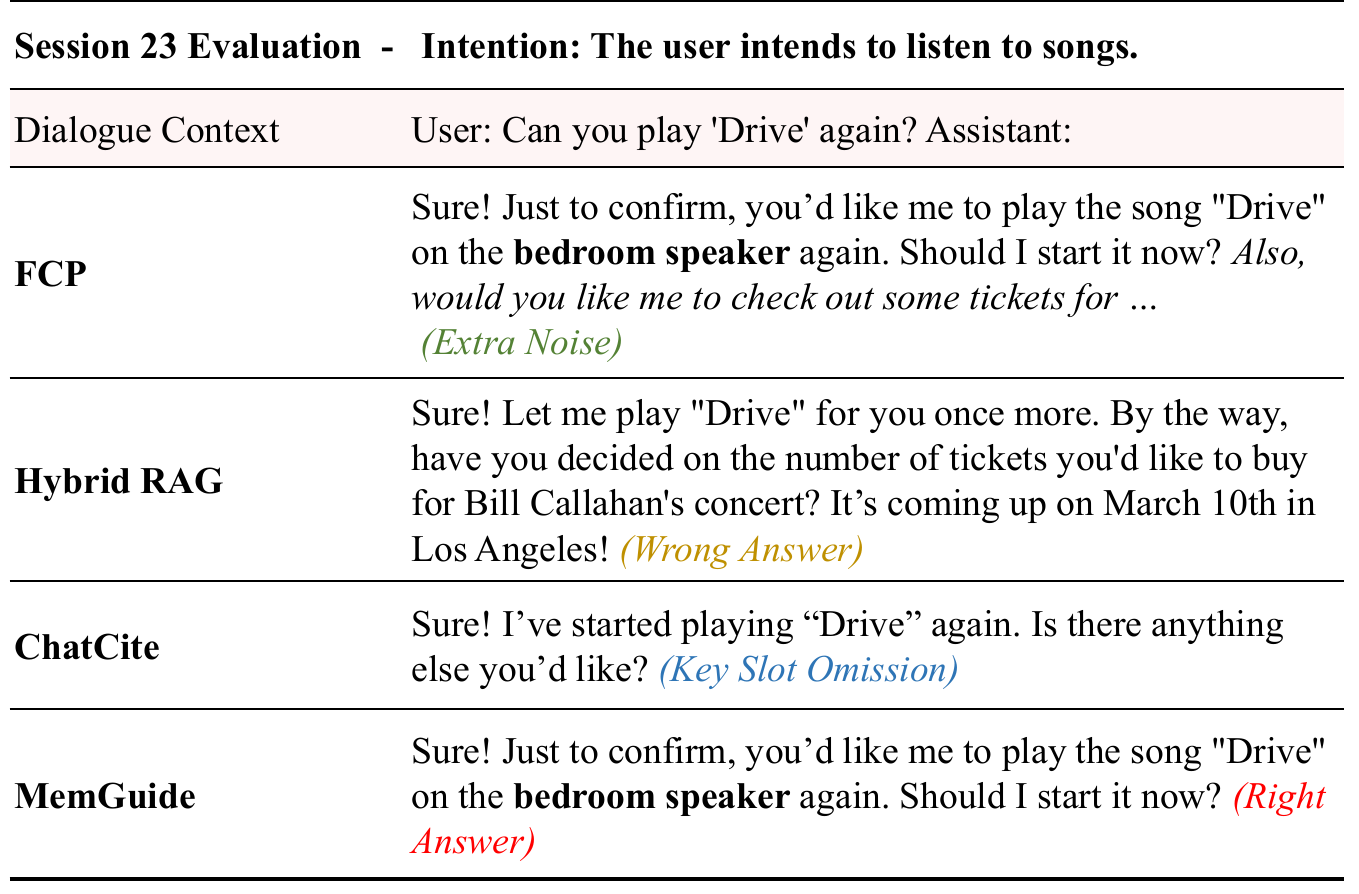}
  \caption{Case study.}
  \label{f-case_study_new}
\end{figure}

\subsection{Case Study}
In our case study, we compared four methods for generating confirmation responses: (1) FCP, (2) Hybrid RAG, (3) ChatCite, and (4) MemGuide. As shown in Figure~\ref{f-case_study_new}, FCP often introduces irrelevant or outdated content due to unfiltered long context, while Hybrid RAG and ChatCite frequently miss critical slot values such as dates or preferences, reflecting retrieval noise or lack of task-specific reasoning. In contrast, MemGuide consistently produces accurate and concise responses by combining intent-aligned retrieval with missing-slot guided filtering. Notably, it achieves higher slot coverage and fluency than other methods, as verified by both GPT-4 scores and manual annotation. These results highlight the value of intent-aware retrieval and task-specific filtering for enhancing response quality in multi-session TOD.
% More details are provided in Appendix \ref{app:case study}.

\section{Conclusion}

We present MemGuide, a two-stage memory-guided framework for multi-session LLM agents. By combining intent-aligned retrieval with missing-slot guided filtering, MemGuide enables task-aware, slot-specific memory selection that surpasses traditional semantic similarity. Evaluated on MS-TOD, our novel benchmark for multi-session TOD, MemGuide significantly improves task success, shortens dialogues, and enhances interaction coherence. These results confirm that structured memory supervision and goal-aware reasoning are critical for developing effective LLM agents.

\bibliography{aaai25}

\newpage
\appendix

\renewcommand{\thefigure}{A.\arabic{figure}}

\section{Appendix A}
\label{app:Promts for dataset generation}

\subsection{Prompt of dialogue generation}
\label{app:dataset_generation_prompt}

\setcounter{figure}{0}  % 从1开始重新编号
We designed a multi-session dialogue prompt (as shown in Figure \ref{fig:Promts for dataset generation}) that generates multi-session dialogue data based on input dialogue intent, task goal, and target session count. Additionally, during the generation process, we annotate whether each utterance is a confirmation response. These annotations, after manual verification, will be used in the main experiment for confirmation-type response generation.

\subsection{Prompt of Task Slot Query Generation}
\label{app:prompt_of_task_slot_query}
During the evaluation process, we design a prompt (as shown in Figure \ref{fig:Promts for Task Slot Querying Generation}) that generates a query corresponding to the missing task attributes based on the current dialogue context and task objectives. The input to this prompt is the dialogue context history and the generated task goal. This query is then used as input to the memory judger to assist in selecting QA memory units that align with the task objectives.

\subsection{Prompts of Confirmation Response Generation}
\label{app:confirmation_generation}
In the evaluation process, we employed a confirmation-type response generation approach to assess the integration performance of multi-session memory in task-oriented dialogues. We designed the prompt as shown in Figure \ref{fig:Promts of Confirmation Response Generation}, which leverages the dialogue context, task objectives, and activated memory units to generate responses.

\subsection{Prompts of GPT4 Evaluation}
\label{app:gpt4_evaluation}
During the evaluation process, we employed a GPT-4 prompt (as shown in Figure \ref{fig:Promts of GPT4 Evaluation}) to assess the quality of confirmation-type responses. This prompt evaluates the response holistically from four perspectives: requirement alignment, content accuracy, language quality, and comparison to the reference answer. The input to this prompt includes the dialogue history, task objectives, the reference response, and the model-generated response. This design ensures that the evaluation of the response is not solely based on the dataset's reference reply but also takes into account multiple factors such as whether the task objectives are met and the overall quality of the response. Such an evaluation approach is more comprehensive.

\begin{table}[t!]
\centering

\resizebox{0.35\textwidth}{!}{%
\begin{tabular}{@{}lc@{}}
\toprule
\textbf{Attribute}            & \textbf{Train} \\ \midrule
Domains                      & 16            \\
Intentions                   & 22            \\
Task goals                   & 4,534         \\
Dialogues                    & 13,441        \\
Utterances                   & 89,152        \\
Avg. slots per task goal     & 4.49          \\ \midrule
Number of individuals         & 565           \\
Avg. intentions per individual & 6.24        \\
Avg. sessions per individual & 23.79         \\
Avg. Utterances per individual & 157.80      \\ \bottomrule
\end{tabular}%
}

\caption{MS-TOD Training Dataset Statistics for Memory Fileter.}
\label{tab:dataset_statistic_train}
\end{table}

\subsection{Prompts of Dialogue State Tracking}
we used a prompt modified from \cite{heck2023chatgptzeroshotdialoguestate} (as shown in Figure \ref{fig:Prompt of Dialogue State Tracking on MultiWOZ 2.2}) that generates the dialogue state for each user turn in the dialogue. Let
\[
A_1 = P \oplus \text{system}:M_1 \oplus \text{user}:U_1
\]
\[
A_t = A_{t-1} \oplus system: M_t \oplus \text{user}:U_t, \quad \forall t \in [2, T]
\]
where P is the task description which provides the model with instructions for how to process a dialogue between a system M and a user U. In contrast to \cite{heck2023chatgptzeroshotdialoguestate}, P does not include the detailed description for slots to challenge ChatGPT's ability to understand the meaning of the slots.
Apart from that, ChatGPT often generated answers with excessively detailed explanations, deviating from the expected response format. To address this issue, a prompt that includes "No explanation!" as an instruction to ChatGPT not to provide detailed explanations was introduced \cite{feng2023llmdrivendialoguestatetracking} and we added this to our prompt.

\newcommand{\cmark}{\ding{51}}  % ✓
\newcommand{\xmark}{\ding{55}}  % ✗
\begin{table*}[t]
\centering
\small
\setlength{\tabcolsep}{5pt}
\renewcommand{\arraystretch}{1.0}
\begin{tabular}{l@{\hskip 6pt}c@{\hskip 6pt}c@{\hskip 6pt}c@{\hskip 6pt}c@{\hskip 6pt}c@{\hskip 6pt}l}
\toprule
\textbf{Dataset} & \textbf{Task Type} & \makecell[c]{Multi-\\Session?} & \makecell[c]{Grounded\\Memory?} & \makecell[c]{User\\Intention?} & \makecell[c]{Retrieval\\Support?} & \makecell[c]{Memory\\Format} \\
\midrule
\textsc{MultiWOZ} \citep{hu2023multi}       & TOD & \xmark & \xmark & \cmark & \xmark & -- \\
\textsc{SGD} \citep{rastogi2020towards}     & TOD & \xmark & \cmark & \cmark & \xmark & schema \\
\textsc{TOAD} \citep{liu-etal-2024-toad}    & TOD & \xmark & \xmark & \cmark & \xmark & -- \\
\textsc{LUCID} \citep{stacey2024lucid}      & TOD & \xmark & \cmark & \cmark & \xmark & latent goal \\
\textsc{MSC} \citep{xu-etal-2022-beyond}    & OD  & \cmark & \cmark & \cmark & \cmark & dialogue history \\
\textsc{CC} \citep{jang-etal-2023-conversation} & OD  & \cmark & \cmark & \xmark & \cmark & persona/dialogue history \\
\textsc{MemoryBank} \citep{zhong2024memorybank} & OD  & \cmark & \cmark & \xmark & \cmark & dialogue history \\
\textsc{LoCoMo} \citep{maharana2024evaluating}  & OD  & \cmark & \cmark & \xmark & \cmark & dialogue history \\
\textsc{LongMemEval} \citep{wu2025longmemeval} & OD  & \cmark & \cmark & \xmark & \cmark & dialogue history \\
\textsc{MS-TOD} (ours)                     & TOD & \cmark & \cmark & \cmark & \cmark & qa memory/dialogue history \\
\bottomrule
\end{tabular}
\caption{Comparison of MS-TOD with representative Task Oriented Dialogue (TOD) and Open Domain (OD) datasets along memory-related attributes.}
\label{tab:dataset_comparison}
\end{table*}

\section{Appendix B}
\label{app:dataset_detail}
\subsection{Training Dataset for Memory Filter}
\label{app:dataset_train}
To ensure that the memory filter generalizes across different domains and scenarios, we generated the training dataset(as shown in Table \ref{tab:dataset_statistic_train}) using the same method described in the main text. The dataset spans 16 domains, 4,534 task goals, and 13,411 dialogues, involving a total of 565 individuals, each with an average of 6.24 intentions. Beyond training the memory filter, this dataset can also serve as an alternative evaluation set for broader benchmarking.

\subsection{Dataset Structure}
\label{app:dataset structure}
MS-TOD encompasses multiple individual task-oriented dialogue datasets, each consisting of several sessions. We present an example of one session (as shown in Figure \ref{fig:mt_tod_session}) from an individual. This session includes a $session\_id$, where a larger value indicates a more recent timestamp. The domain represents the specific field or area of the dialogue. The $reference\_dialogue
\_id$ corresponds to the $dialogue\_id$ in the original SGD dataset that shares the same task objective. The $exist\_confirmation$ indicates whether the session contains a confirmation-type response and whether it is an evaluation target. The intent represents the specific purpose or goal of the dialogue. The content stores the actual dialogue text. The $task\_goal$ includes task slots and their corresponding attribute values. Each individual contains dozens of session data structured as described above.

\subsection{Human Validation Protocol}
\label{app:human_validation}
To ensure the realism, coherence, and usability of MS-TOD, we apply a structured human validation process during dataset construction. This process involves three research assistants with prior experience in natural language processing and dialogue systems. The validation pipeline includes the following stages:

\begin{enumerate}
    \item \textbf{Intent and Slot Accuracy Check.} For each dialogue turn derived from the SGD intent annotations, annotators verify whether the intent is preserved and whether all required slot values are present and semantically correct.
    
    \item \textbf{Redundancy Removal.} Annotators manually review and remove multi-session dialogues that contain excessive repetition across sessions, which could undermine diversity and realism.
    
    \item \textbf{Confirmation Accuracy Validation.} For final-session confirmation-type utterances, annotators examine whether the confirmed slot values align with the task goal. Mismatched, ambiguous, or hallucinated confirmations are flagged and discarded.
    
    \item \textbf{Dialogue Coherence Filtering.} Dialogues that fail to complete any defined task goal are considered incoherent. Sessions missing necessary confirmation-type turns are also excluded to ensure logical task flow.
    
    \item \textbf{Intent Redundancy Filtering.} Episodes exhibiting unnatural repetition of similar intents across turns or sessions are excluded, as such patterns deviate from realistic multi-session user behavior.
\end{enumerate}

This multi-stage quality control procedure yields a filtered evaluation subset used for system benchmarking. The validation process ensures that the dataset aligns with realistic task-oriented dialogue patterns and supports the evaluation of multi-session memory-aware dialogue systems.

\subsection{Intent-driven QA Memory}
\label{app:qa memory}
For each historical session, we generated an intent description and the corresponding QA memory (as shown in Figure \ref{fig:mt_tod_qa_summary}) for the objectives of that intent description. The QA memory consists of multiple QA pairs, where each query is a question about a task attribute under that intent, and the answer is the slot value corresponding to that task attribute.

\subsection{Dataset Design Rationale}
\label{sec:dataset_design}
\paragraph{Choice of Seed Dataset.} We select the Schema-Guided Dialogue (SGD) dataset as the foundation for constructing MS-TOD. Compared to other popular benchmarks like MultiWOZ, SGD provides broader domain coverage, a larger and more diverse set of user intents, and a schema-driven annotation format that supports extensibility and dynamic intent representation. These characteristics make SGD more suitable for modeling realistic, multi-domain, and multi-session interactions. A detailed comparison is shown in Table~\ref{tab:multiwoz_sgd_compare}.

\paragraph{Design of Memory Bank Structure.}
Each MS-TOD memory bank contains 20 sessions involving more than six distinct user intents. This structure is informed by two factors. First, prior multi-session datasets such as LoCoMo \citep{} typically use memory segments with 20+ sessions, providing a reference for session scale under long-term memory settings. Second, based on analysis of the SGD schema, each user intent generally corresponds to fewer than 10 slot types. Organizing ~3 sessions per intent enables natural progression while minimizing redundancy. As a result, grouping 6–8 distinct intents yields a total of around 20 sessions per memory bank, balancing diversity, realism, and memory demand.

\begin{table*}[t!]
\centering
\small
\begin{tabular}{lcc}
\toprule
\textbf{Dimension} & \textbf{MultiWOZ} & \textbf{SGD} \\
\midrule
\# Domains & 7 & 20 \\
Avg. Intents per Domain & 8–10 & 10–15 \\
Total Intents & $\sim$60 & $\sim$200 \\
Annotation Structure & Fixed, manually updated & Schema-driven, extensible \\
Cross-Domain Intent Interaction & Limited (2–3 domain combos) & Rich (multi-domain intent chains) \\
\bottomrule
\end{tabular}
\caption{Comparison between MultiWOZ and SGD datasets.}
\label{tab:multiwoz_sgd_compare}
\end{table*}

\section{Appendix C}

\subsection{Memory Retrieval Comparision}
Table~\ref{tab:activation_modules} compares the performance of different retrieval methods. \textbf{text-embed3-small} achieves the highest recall across all thresholds, with 0.702 at Recall@3, 0.792 at Recall@5, and 0.905 at Recall@10, demonstrating superior retrieval capability. Among other models, \textbf{nv-embed-v2} and \textbf{bge-large-en-v1.5} also perform well, while traditional retrieval methods like \textbf{BM25} remain competitive at Recall@10 but lag behind embedding-based methods at lower recall levels. \textbf{T5-base} and \textbf{BERT-based models} exhibit lower recall, suggesting that general pre-trained models are less effective for specialized memory retrieval. These results highlight \textbf{text-embed3-small} as the most effective choice for long-term memory activation in multi-session dialogues.

\subsection{Effectiveness of the Proactive Response Strategy}
\label{app:proactive_response_effect}
To better understand the impact of the proactive response strategy, we present a complementary analysis that examines two distinct metrics: slot accuracy measured during the confirmation phase and the final task success rate. Although these metrics reflect different aspects of system performance—localized slot-level correctness versus overall goal completion—they jointly capture the effectiveness of proactive correction.

As shown in Table~\ref{tab:proactive_response_effect}, slot accuracy remains relatively low (ranging from 0.48 to 0.62) before correction, indicating frequent omission or mismatch in predicted slot values. Nevertheless, the final task success rates reach 0.87 or higher across all models after proactive correction is applied. This pattern suggests that the proactive response strategy plays a critical role in bridging the gap between partial slot-level understanding and complete task execution by enabling the system to recover from intermediate errors through user interaction.

\begin{table}[t!]
\centering
\small
\renewcommand{\arraystretch}{1.1}
\setlength{\tabcolsep}{5pt}
\begin{tabular}{lcc}
\toprule
\textbf{Model} & \textbf{Slot Acc. (Pre)} & \textbf{Task Rate (Post)} \\
\midrule
LLaMA3-8B     & 0.62 & 0.92 \\
Qwen-7B       & 0.48 & 0.87 \\
Mistral-7B    & 0.59 & 1.00 \\
GPT4o-mini    & 0.61 & 0.99 \\
\bottomrule
\end{tabular}
\caption{Effectiveness of the Proactive Response Strategy. Slot accuracy is measured before correction, and task rate reflects the final success after proactive clarification.}
\label{tab:proactive_response_effect}
\end{table}

\subsection{Human Evaluation Details}
\label{app: human_evalution}
Table \ref{tab:human_eval} presents the results of human evaluation, including accuracy, informativeness, and coherency scores. Accuracy is rated on a scale of 0 to 1, while informativeness and coherency are rated from 0 to 3. The average scores in Table \ref{tab:combined_results} are computed using a weighted sum with weights of 1, 1/3, and 1/3. All evaluations were conducted in a blind review manner to compare the response quality of FCP and MemGuide. Additionally, the Confirmation-type Response type assesses the response quality after memory-guided dialogue planning, while the multi-turn evaluation focuses on dialogues under the proactive response strategy, continuing until task completion or forced termination.

\begin{table}[t!]
\centering
\resizebox{0.48\textwidth}{!}{% % 调整宽度为 0.4
\begin{tabular}{l S[table-format=1.3] S[table-format=1.3] S[table-format=1.3]}
\toprule
\textbf{Activation Module} & \textbf{Recall@3} & \textbf{Recall@5} & \textbf{Recall@10} \\
\midrule
bm25                     & 0.642 & 0.721 & 0.842 \\
t5-base                  & 0.443 & 0.575 & 0.773 \\
bert-base                & 0.463 & 0.584 & 0.785 \\
bert-large               & 0.401 & 0.530 & 0.730 \\
nv-embed-v2              & 0.668 & 0.769 & 0.896 \\
bge-large-en-v1.5        & 0.681 & 0.761 & 0.888 \\
\textbf{text-embed3-small} & \textbf{0.702} & \textbf{0.792} & \textbf{0.905} \\
\bottomrule
\end{tabular}%
}

\caption{Performance evaluation of activation modules on memory retrieval}
\label{tab:activation_modules}
\end{table}

\begin{table}[t!]
\centering
\small
% 模仿第一个表格设置
\renewcommand{\arraystretch}{1.3} % 调整行间距
\setlength{\tabcolsep}{2pt}       % 调整列间距
\fontsize{9pt}{11pt}\selectfont   % 设置字体大小
\begin{tabular}{llccc}
\toprule
\textbf{Model} & \textbf{Setting} & \textbf{Slot Accuracy} & \textbf{BLEU} & \textbf{ROUGE} \\
\midrule
\multirow{2}{*}{\textbf{LLaMA3-8B}} 
    & FCP & \textbf{0.62} & \textbf{10.47} & 28.59 \\
    & \cellcolor[gray]{0.9} MemGuide 
      & \cellcolor[gray]{0.9}0.56 
      & \cellcolor[gray]{0.9}9.86   
      & \cellcolor[gray]{0.9}\textbf{30.39} \\ 
\hdashline
\multirow{2}{*}{\textbf{Qwen-7B}}
    & FCP & 0.48 & 10.33 & 29.77 \\
    & \cellcolor[gray]{0.9} MemGuide 
      & \cellcolor[gray]{0.9}\textbf{0.55} 
      & \cellcolor[gray]{0.9}\textbf{10.90} 
      & \cellcolor[gray]{0.9}\textbf{31.28} \\ 
\hdashline
\multirow{2}{*}{\textbf{Mistral-7B}}
    & FCP & 0.59 & \textbf{10.09} & \textbf{28.42} \\
    & \cellcolor[gray]{0.9} MemGuide 
      & \cellcolor[gray]{0.9}\textbf{0.56} 
      & \cellcolor[gray]{0.9}6.66  
      & \cellcolor[gray]{0.9}24.64 \\ 
\hdashline
\multirow{2}{*}{\textbf{GPT4o-mini }}
    & FCP & 0.61 & \textbf{20.30} & \textbf{43.49} \\
    & \cellcolor[gray]{0.9} MemGuide 
      & \cellcolor[gray]{0.9}\textbf{0.68} 
      & \cellcolor[gray]{0.9}13.6  
      & \cellcolor[gray]{0.9}35.20 \\
\bottomrule
\end{tabular}
\caption{Performance comparison of task-oriented dialogue models with and without long-term memory integration: Slot Accuracy, BLEU, and ROUGE metrics.}
\label{tab:slot_bleu_rouge}
\end{table}

\begin{table}[t!]
\centering
\renewcommand{\arraystretch}{1.2}  % 调整表格行间距
\setlength{\tabcolsep}{4pt}       % 调整列间距
\fontsize{9pt}{11pt}\selectfont   % 调整字体大小
\begin{tabular}{lccc}
\toprule
\textbf{Model} & \textbf{Slot Accuracy} & \textbf{BLEU} & \textbf{ROUGE} \\
\midrule
AutoTOD  & 0.61 & 3.34 & 24.07 \\
MemGuide     & \textbf{0.68} & \textbf{5.47} & \textbf{25.03} \\
\bottomrule
\end{tabular}
\caption{Performance comparison on Slot Accuracy, BLEU, and ROUGE.}
\label{tab:tod_slot_bleu_rouge}
\end{table}

\begin{table*}[t!]
\centering
\small
\renewcommand{\arraystretch}{1.3} % 调整行间距
\begin{tabular}{llcccccc}
\toprule
\textbf{Model} & \textbf{Setting} & \multicolumn{3}{c}{\textbf{Confirmation-type Response}} & \multicolumn{3}{c}{\textbf{Multi-Turn}} \\
\cmidrule(lr){3-5} \cmidrule(lr){6-8}
 &  & \textbf{Accuracy} & \textbf{Informativeness} & \textbf{Coherency} & \textbf{Accuracy} & \textbf{Informativeness} & \textbf{Coherency} \\
\midrule
\multirow{2}{*}{\textbf{GPT4o-mini}} 
    & FCP & 0.62 & 1.83 & 1.90 & 0.81 & 1.92 & 2.44 \\
    & \cellcolor[gray]{0.9} MemGuide & \cellcolor[gray]{0.9}\textbf{0.65} & \cellcolor[gray]{0.9}\textbf{2.38} & \cellcolor[gray]{0.9}\textbf{2.48} & \cellcolor[gray]{0.9}\textbf{0.87} & \cellcolor[gray]{0.9}\textbf{1.93} & \cellcolor[gray]{0.9}\textbf{2.74} \\ \hdashline
\multirow{2}{*}{\textbf{LLaMA}} 
    & FCP & 0.56 & 1.47 & 1.74 & 0.78 & 1.64 & 2.36 \\
    & \cellcolor[gray]{0.9} MemGuide & \cellcolor[gray]{0.9}\textbf{0.61} & \cellcolor[gray]{0.9}\textbf{1.98} & \cellcolor[gray]{0.9}\textbf{2.16} & \cellcolor[gray]{0.9}\textbf{0.88} & \cellcolor[gray]{0.9}\textbf{2.51} & \cellcolor[gray]{0.9}\textbf{2.71} \\ \hdashline
\multirow{2}{*}{\textbf{Qwen}} 
    & FCP & 0.43 & 1.24 & 1.85 & 0.82 & 1.60 & 2.02 \\
    & \cellcolor[gray]{0.9} MemGuide & \cellcolor[gray]{0.9}\textbf{0.54} & \cellcolor[gray]{0.9}\textbf{1.70} & \cellcolor[gray]{0.9}\textbf{2.30} & \cellcolor[gray]{0.9}\textbf{0.92} & \cellcolor[gray]{0.9}\textbf{1.93} & \cellcolor[gray]{0.9}\textbf{2.47} \\ \hdashline
\multirow{2}{*}{\textbf{Mistral}} 
    & FCP & 0.58 & 1.63 & 1.99 & 0.89 & 2.49 & 2.72 \\
    & \cellcolor[gray]{0.9} MemGuide & \cellcolor[gray]{0.9}\textbf{0.61} & \cellcolor[gray]{0.9}\textbf{2.06} & \cellcolor[gray]{0.9}\textbf{2.08} & \cellcolor[gray]{0.9}\textbf{0.93} & \cellcolor[gray]{0.9}\textbf{2.74} & \cellcolor[gray]{0.9}\textbf{2.85} \\
\bottomrule
\end{tabular}
\caption{Comparison of different models on human evaluation metrics: accuracy, informativeness, and coherence. The results are presented for both confirmation-type responses and multi-turn dialogue settings, comparing standard inference (`FCP`) with memory-augmented processing (`MemGuide`).}
\label{tab:human_eval}
\end{table*}

\subsection{Additional Evaluation Metrics}
Table~\ref{tab:slot_bleu_rouge} compares the performance of task-oriented dialogue models with and without memory-augmented processing (MemGuide) across Slot Accuracy, BLEU, and ROUGE metrics. The results reveal a trade-off between structured slot accuracy and response fluency. In most models, MemGuide slightly reduces slot accuracy, as seen in LLaMA3-8B, which drops from 0.62 to 0.56, and Mistral-7B, which decreases from 0.59 to 0.56. However, GPT4o-mini benefits from MemGuide, achieving the highest slot accuracy of 0.68. BLEU scores generally decline, suggesting that MemGuide shifts responses away from verbatim accuracy towards greater contextual adaptability. Mistral-7B drops from 10.90 to 6.66, and LLaMA3-8B decreases from 10.47 to 9.86. Conversely, ROUGE scores improve with MemGuide in several cases. LLaMA3-8B increases from 28.59 to 30.39, and Qwen-7B rises from 29.77 to 31.28, indicating enhanced informativeness and coherence. However, Mistral-7B experiences a slight decrease in ROUGE from 28.42 to 24.64. Overall, the results suggest that MemGuide enhances response informativeness while slightly compromising slot accuracy and BLEU, highlighting a trade-off between structured information retention and more natural, contextually aware responses.

Table~\ref{tab:tod_slot_bleu_rouge} presents the performance comparison between AutoTOD and MemGuide on Slot Accuracy, BLEU, and ROUGE. The results indicate that MemGuide consistently outperforms AutoTOD across all three metrics, demonstrating its effectiveness in enhancing dialogue quality. Slot Accuracy improves from 0.61 to 0.68, indicating better tracking of task-specific information. BLEU increases from 3.34 to 5.47, reflecting more precise and fluent responses. ROUGE also shows a slight improvement, rising from 24.07 to 25.03, suggesting that MemGuide enhances informativeness and coherence. These results highlight the advantages of memory-augmented processing, which enables more accurate and contextually relevant dialogue generation.

\section{Appendix D}
\label{app:case study}

\subsection{Multi-session Dialogue Context Comparison}
Figure \ref{f-case_study} presents four different configurations of conversation contexts not shown in the main paper. Specifically, (1) Full conversation history includes every session from the dialogue history as prompt input to the reader. (2) Retrieval-based methods retrieve the dialogue sessions most relevant to the current session (Session 23) and append them to the reader’s context (3) Retrieving a summary compiles a summary of past sessions (Sessions 1 to 22) for inclusion alongside the current context. Finally, (4) MemGuide integrates QA memory with the Session 23 context to generate responses. By illustrating these detailed contexts, Figure \ref{f-case_study} provides further insights into how each approach manages multi-session dialogue.

\subsection{MemGuide vs. RAG}
To better understand how CoT reasoning and memory reranking affect confirmation response generation, we present a step-by-step case study comparing MemGuide and standard RAG (Appendix Table~\ref{tab:cot_rag_case_study}). In this example, the user attempts to confirm a restaurant reservation. While both systems retrieve similar QA memory candidates, the standard RAG model fails to detect missing slot information (e.g., number of people), resulting in an incomplete and partially inaccurate response. In contrast, MemGuide use Chain-of-Thought explicitly identifies missing task information (e.g., time, headcount) through reasoning, refines the retrieved memories via the Memory Judger, and generates a more complete and contextually appropriate confirmation. This illustrates how structured reasoning and selective memory grounding improve slot coverage and reduce factual errors in multi-turn dialogue.

% \begin{figure}[t!]
%   \centering
%   \includegraphics[width=\linewidth]{img/memory_activation_judger_top_5.pdf}
%   \caption{Impact of memory judger on memory activation performance across different embedding models.}
%   \label{f-ablation_study}
% \end{figure}

\newpage
\begin{figure*}[t!]
\begin{tcolorbox}[
colframe=black!75!white, 
colback=white, sharp corners, 
boxrule=0.8pt, width=\textwidth,
title=Prompts of the Dataset Generation
] 
User Prompt:

"""\\
Help me generate an English conversation under the \{dialogue\_intent\} intent,
      where \{task\_goal\}. The conversation should be between a user and an assistant, 
      and it should be split into \{task\_goal\_length\} sessions at different points in time, 
      with continuity and connection between the sessions and each session should not less than 6 turns.
      Additionally, the final session must include a assistant response containing a complete confirmation-type utterance before the user confirms, and this utterance should be marked with `is\_confirmation` set to `True`.
      and the user must provide a final confirmation response at the end of the final session. 
      For all other sessions, the conversation should end with an assistant's polite declarative statement. 

"""

System Prompt:

"""
You are dialogue generator assistant. \\
The sessions should be clearly separated, and the conversation should be formatted as follows:\\
Each turn should be a dictionary entry.\\
The conversation should be in the format of a list of sessions, where each session is a list of dictionaries representing each turn.\\
Each dictionary entry should have two keys: speaker (either 'user' or 'assistant') and text (the spoken dialogue).\\
Except for final session, each session should be a seperate dialogue and include a complete dialogue structure, beginning with a greeting from the user and ending with an assistant's polite declarative statement.\\
Feel free to expand the dialogue with additional relevant details, but avoid redundant expressions or repeating the same phrases.\\
Reponse me with a json format\\
\begin{lstlisting}
{
    "sessions": [
        [
            {
                "speaker": "xx",
                "text": "xx"
            },
            {
                "speaker": "xx",
                "text": "xx"
            }
        ]
    ]
}
\end{lstlisting}
"""

\end{tcolorbox} 
\caption{Prompts of the Dataset Generation}
\label{fig:Promts for dataset generation}
\end{figure*}

\newpage

\begin{figure*}
\begin{tcolorbox}[
colframe=black!75!white, 
colback=white, sharp corners, 
boxrule=0.8pt, width=\textwidth,
title=Prompts of the Task Slot Querying Generation
] 
"""\\
Please help me generate questions, 
    based on the provided \{conversation history\}, that correspond to unanswered attributes in the task goal \{task\_attributes\}. \\
    1. The questions should start with 'What,' 'When,' 'Why,' 'How,' or 'Where.' \\
    2. Ensure that the generated questions are in third person.\\
    fill the following json:
            \{
            [Question],
            \}\\
"""

\end{tcolorbox} 
\caption{Prompts of the Task Slot Querying Generation}
\label{fig:Promts for Task Slot Querying Generation}
\end{figure*}

\begin{figure*}
\begin{tcolorbox}[
colframe=black!75!white, 
colback=white, sharp corners, 
boxrule=0.8pt, width=\textwidth,
title=Prompts of Confirmation Response Generation
] 
"""
You are an dialogue assistant. \\
Generate a confirmation response based on the user\'s utterance. Include any relevant task goals [TASK GOALS] identified in the dialogue or related memory [MEMORY]. If [MEMORY] is unavailable, construct your response accurately and comprehensively using the provided conversation details. Ensure your reply acknowledges the user\'s request clearly and incorporates relevant information from both the dialogue and the related memory units [MEMORY]. \\
$[$TASK GOAL$]$ \\
\{task\_goal\} \\

[MEMORY] \\
\{memory\} \\
"""

\end{tcolorbox} 
\caption{Prompt of Confirmation Response Generation}
\label{fig:Promts of Confirmation Response Generation}
\end{figure*}

\begin{figure*}
\begin{tcolorbox}[
colframe=black!75!white, 
colback=white, sharp corners, 
boxrule=0.8pt, width=\textwidth,
title=Prompt of Dialogue State Tracking on MultiWOZ 2.2
] 
"""Consider the following list of concepts, called "slots" provided to you as a json list.\\

"slots": \{\\
    "attraction-area",\\
    "attraction-name",\\
    "attraction-type",\\
    "bus-day",\\
    "bus-departure",\\
    "bus-destination",\\
    "bus-leaveat",\\
    "hospital-department",\\
    "hotel-area",\\
    "hotel-bookday",\\
    "hotel-bookpeople",\\
    "hotel-bookstay",\\
    "hotel-internet",\\
    "hotel-name",\\
    "hotel-parking",\\
    "hotel-pricerange",\\
    "hotel-stars",\\
    "hotel-type",\\
    "restaurant-area",\\
    "restaurant-bookday",\\
    "restaurant-bookpeople",\\
    "restaurant-booktime",\\
    "restaurant-food",\\
    "restaurant-name",\\
    "restaurant-pricerange",\\
    "taxi-arriveby",\\
    "taxi-departure",\\
    "taxi-destination",\\
    "taxi-leaveat",\\
    "train-arriveby",\\
    "train-bookpeople",\\
    "train-day",\\
    "train-departure",\\
    "train-destination",\\
    "train-leaveat",\\
\}\\

Now consider the following dialogue between two parties called the "system" and "user". Can you tell me which of the "slots" were updated by the "user" in its latest response to the "system"? Present the updates in JSON format. If no "slots" were updated, return an empty JSON list. If you encounter "slots" that were requested by the "user" then fill them with "?". If the user informed that he did not care about a "slot", fill it with "dontcare". Return the output in JSON format and no explanation!\\
\{dialogue\}\\
"""

\end{tcolorbox} 
\caption{Prompt of Dialogue State Tracking on MultiWOZ 2.2}
\label{fig:Prompt of Dialogue State Tracking on MultiWOZ 2.2}
\end{figure*}

\begin{figure*}
\begin{tcolorbox}[
colframe=black!75!white, 
colback=white, sharp corners, 
boxrule=0.8pt, width=\textwidth,
title=Prompts of GPT4 Evaluation
] 
"""
You are a strict and objective evaluator. Your task is to assess the quality of the final predicted response using the provided conversation context, the user’s target goal attributes, and a reference answer. Your evaluation should be fair, professional, and reflect an expert judgment of the response’s quality.

[Dialogue Context] \\
$\{\{conversation_history\}\}$

[Task Goal] \\
$\{\{task\_goal\}\}$

[reference\_answer] \\
$\{\{reference\_anwser\}\}$

[predict\_answer] \\
$\{\{predict\_answer\}\}$

Evaluation Criteria:\\
Requirement Alignment: Does the final predict\_answer meet the user’s task goal? \\
Content Accuracy: Is the information in the final response correct, clear, and logically organized? \\
Language Quality: Is the language fluent, coherent, and readable? Are there any obvious grammatical or word choice errors? \\
Comparison to Reference Answer: Compared to the reference answer, how does the final response differ in terms of completeness, professionalism, and clarity?\\
Overall Score: Assign a score from 1 to 10 (10 being the best), considering all of the above factors. \\

The evaluation must be structured in the following JSON format: \\
```json \\
\{ \\
  "Rationale": "<Explain the rationale of your score.>", \\
  "Score": <An integer score from 1 to 10.> \\
\} \\
```
"""

\end{tcolorbox} 
\caption{Prompts of GPT4 Evaluation}
\label{fig:Promts of GPT4 Evaluation}
\end{figure*}

% 定义 JSON 语法高亮
\lstdefinelanguage{json}{
    basicstyle=\ttfamily\small
}

\begin{figure*}
\begin{tcolorbox}[
colframe=black!75!white, 
colback=white, sharp corners, 
boxrule=0.8pt, width=\textwidth,
title=MS-TOD dialogue session structure
] 
\begin{lstlisting}[language=json, breaklines=true]
{
  "session_id": 9,
  "domain": "Travel",
  "reference_dialogue_id": "66_00101",
  "exist_confirmation": true,
  "intent": "ReserveHotel",
  "content": [
    {
      "speaker": "user",
      "utterance": "Hi again, I'm ready to finalize the booking for Aloft Portland Airport At Cascade Station.",
      "is_confirmation": false
    },
    {
      "speaker": "assistant",
      "utterance": "Just to confirm, you are booking 1 room at Aloft Portland Airport At Cascade Station, Portland, from March 5th to March 7th. The room is a standard king room with free Wi-Fi and a 24-hour cancellation policy. Is that correct?",
      "is_confirmation": true
    },
    {
      "speaker": "user",
      "utterance": "Yes, that is correct.",
      "is_confirmation": false
    },
    {
      "speaker": "assistant",
      "utterance": "Excellent! Your room has been successfully booked. You will receive a confirmation email shortly.",
      "is_confirmation": false
    },
    {
      "speaker": "user",
      "utterance": "Thank you so much for your help!",
      "is_confirmation": false
    },
    {
      "speaker": "assistant",
      "utterance": "You're welcome! Have a great stay in Portland.",
      "is_confirmation": false
    }
  ],
  "task_goal": {
    "hotel_name": "Aloft Portland Airport At Cascade Station",
    "location": "Portland",
    "check_in_date": "March 5th",
    "check_out_date": "March 7th",
    "number_of_rooms": 1
  }
}
\end{lstlisting}

\end{tcolorbox} 
\caption{MS-TOD Session Structure.}
\label{fig:mt_tod_session}
\end{figure*}

\begin{figure*}
\begin{tcolorbox}[
colframe=black!75!white, 
colback=white, sharp corners, 
boxrule=0.8pt, width=\textwidth,
title=MS-TOD Intent Description and QA Memory
] 
\begin{lstlisting}[language=json, breaklines=true]
{
  "9": {
    "intent_description": "The user's intent is to finalize and confirm a hotel booking for a specific room at Aloft Portland Airport At Cascade Station, including details about the stay dates and room type.",
    "qa_summary": [
      {
        "Question": "What type of room did the user book?",
        "Answer": "The user booked a standard king room."
      },
      {
        "Question": "When is the user's reservation?",
        "Answer": "The user's reservation is from March 5th to March 7th."
      },
      {
        "Question": "Where is the user's reservation located?",
        "Answer": "The user's reservation is located at Aloft Portland Airport At Cascade Station."
      },
      {
        "Question": "What amenities are included in the user's reservation?",
        "Answer": "The user's reservation includes free Wi-Fi."
      },
      {
        "Question": "What is the cancellation policy for the user's booking?",
        "Answer": "The cancellation policy for the user's booking is 24 hours."
      }
    ]
  }
}
\end{lstlisting}

\end{tcolorbox} 
\caption{Intent description and QA Memory in MT-TOD.}
\label{fig:mt_tod_qa_summary}
\end{figure*}
\newpage

\begin{table*}[h]
\centering
\small
\renewcommand{\arraystretch}{1.2}
\begin{tabular}{|p{3.4cm}|p{6.2cm}|p{6.2cm}|}
\hline
\textbf{Process} & \textbf{MemGuide} & \textbf{RAG} \\
\hline
\multicolumn{3}{|c|}{\textbf{Input and Intent}} \\
\hline
Dialogue History & \textit{User: Have you completed the reservation at Gen Korean BBQ House?} & \textit{User: Have you completed the reservation at Gen Korean BBQ House?} \\
Intention Description & The user wants to confirm restaurant reservation. & The user wants to confirm restaurant reservation. \\
\hline
\multicolumn{3}{|c|}{\textbf{Memory Retrieval}} \\
\hline
Retrieved QA Memory Candidates & \multicolumn{2}{p{12.3cm}|}{
\textbf{Rank 1:}
\begin{itemize}
\item Q: What is the time of the reservation? A: March 1st
\item Q: What is the address of the reservation? A: Los Angeles
\end{itemize}
\textbf{Rank 2:}
\begin{itemize}
\item Q: What is the time of reservation? A: March 4th
\item Q: How many people are there? A: 2
\item Q: What is the address of the restaurant? A: Gen Korean BBQ House in Milpitas
\item Q: What is the time of reservation? A: March 1st
\item Q: What is the address of the restaurant? A: Gen Korean BBQ House in Milpitas
\end{itemize}
} \\
\hline
\multicolumn{3}{|c|}{\textbf{CoT Reasoning}} \\
\hline
Task Goal & Reserve Restaurant & --- \\
\hline
Missing Slots & Time, Number of people & --- \\
\hline
Missing Query & When is the time of reservation? How many people are there? & --- \\
\hline
\multicolumn{3}{|c|}{\textbf{Memory Filter (Reranking)}} \\
\hline
Reranked Memory Units &
\begin{itemize}
\item Q: What is the time of reservation? A: March 4th
\item Q: How many people are there? A: 2
\item Q: What is the address of the restaurant? A: Gen Korean BBQ House in Milpitas
\end{itemize}
& Same as retrieved \\
\hline
\multicolumn{3}{|c|}{\textbf{Refinement-grounded Response Generation}} \\
\hline
Confirmation Response &
Just to confirm, it's a reservation for 2 at Gen Korean BBQ House in Milpitas on March 4th at 12:15 pm, with a request for a quieter table. Is that correct?
&
To confirm, it's a reservation for 2 at Gen Korean BBQ House Los Angeles on March 1st. Is that correct? \\
\hline
\end{tabular}
\caption{Step-by-step comparison of MemGuide vs. standard RAG in confirmation response generation.}
\label{tab:cot_rag_case_study}
\end{table*}

\newpage

% \begin{figure*}[t!]
%   \centering
%   \includegraphics[width=\linewidth]{img/case_study_for_TOD.pdf}
%   \caption{Comparison of confirmation response generation across four approaches: (1) Direct Prompting with the full conversation history, (2) Hybrid RAG retrieving relevant dialogue history, (3) Hybrid RAG retrieving a summary of the conversation, and (4) MemGuide with intention-based QA memory.}
%   \label{f-case_study}
%   % \vspace{-10pt}
% \end{figure*}

\end{document}